\documentclass[11pt]{article}

\usepackage{acl}

\usepackage{times}
\usepackage{latexsym}
\usepackage[T1]{fontenc}
\usepackage[utf8]{inputenc}
\usepackage{microtype}
\usepackage{inconsolata}
\usepackage{graphicx}
\usepackage{booktabs}
\usepackage{multirow}
\usepackage{array}
\usepackage{amsmath,amssymb}
\usepackage{xcolor}
\usepackage{listings}
\graphicspath{{figures/}}

\definecolor{codebg}{rgb}{0.97,0.97,0.97}
\newcommand{\depth}{\ensuremath{\mathrm{depth}}}
\newcommand{\code}[1]{\texttt{#1}}
\newcommand{\famname}[1]{\textsf{#1}}
\newcommand{\sysLethe}{\textsc{Lethe}}
\newcommand{\sysMem}{\textsc{Mem0}}
\newcommand{\sysPalace}{\textsc{MemPalace}}
\newcommand{\sysAmem}{\textsc{A-MEM}}
\newcommand{\sysForget}{\textsc{ForgetEval}}

\title{Control-Plane Placement Shapes Forgetting: An Architectural Study of Agent Memory Across Thirteen System Configurations}

\author{Dongxu Yang\thanks{Correspondence: \texttt{wayland0916@gmail.com}} \\
  DeepLethe}

\begin{document}
\maketitle

\begin{abstract}
\emph{Where} an LLM sits in an agent memory pipeline ---
between the \emph{recall plane} that retrieves stored facts
(extensively benchmarked) and the \emph{control plane} that
mutates them via \code{supersede}, \code{release}, \code{purge}
(largely untested) --- shapes which forgetting failure modes
the system recovers.  Comparing
thirteen system configurations on a common 385-case adversarial
surface (four deterministic, two vec-only, two inscribe-LLM,
two KG-abstraction, one inscribe+mutation joint, two
mutation-LLM hook backends) --- we observe three placement
regimes with
\emph{partly complementary} coverage: deterministic primitives
suffice for lexical/temporal categories but fail canonicalization
(${\leq}5\,\%$ on identifier-obfuscation, 0\,\% on
cross-lingual); inscribe-time LLM recovers canonicalization
(100\,\%) but cannot help intent-aware deletion (0\,\% on
prefix-collision and compound-fact); a mutation-time hook
recovers intent-aware deletion (78--85\,\%) and brightens
nearly all categories simultaneously: a \textbf{capability
lift of $+$22.6 to $+$24.1\,pt} on the 345
non-primitive-existence cases (deterministic
70.1--70.7\,\% ${\to}$ hook 93.3--94.2\,\%), plus
${\sim}{+}6$\,pt from edit-primitive availability on
\famname{compound\_fact} (headline overall
91.7--93.2\,\%); ${\sim}\$0.17$ per 385-case run, mutation
latency ${\sim}2.3$\,s/case vs.\ 64--191\,ms/case
deterministic, recall hot path unchanged.

We expose the placement trade-off via \textbf{\sysForget{}}, a
1000-case templated suite plus a 385-case adversarial layer
(132 hand-crafted core + 253 LLM-drafted oracle-validated)
scored by deterministic substring match, paired with a
six-method \code{Adapter} Protocol with honest \textbf{N/A
scoring} that lets heterogeneous memory stores enter the
evaluation in ${\sim}130$ lines.  Admission is corroborated
by 10-annotator IAA (Fleiss' $\kappa = 0.958$) and a 77-case
external-authored subset (four blind contributors) that
replicates the canonicalization asymmetry and amplifies the
joint-placement lift ($+27.8$\,pt).  Production
failures are predominantly forgetting failures (rotated
credentials still recommended, GDPR-deleted records still
surfacing) rather than recall failures, yet existing memory
benchmarks measure only recall.  \sysForget{} and all adapters
are released under MIT.
\end{abstract}

\section{Introduction}

An agent's memory has two control surfaces: a \emph{recall plane}
that retrieves stored facts, and a \emph{control plane} that
mutates them --- supersede, release, purge.  This paper's claim
is that \emph{where} an LLM sits relative to these two planes ---
not \emph{whether} one is present --- determines which forgetting
failures an agent can recover from.  The field has saturated the
recall plane and left the control plane untested: every memory
framework \cite{mem0,memgpt,letta,zep,hipporag,amem,cognee}
races to recall harder and never lose a fact, and the benchmarks
follow --- LongMemEval~\cite{longmemeval} scores R@$k$,
MTEB~\cite{mteb} and BEIR~\cite{beir} measure stateless retrieval,
LOCOMO scores conversational recall with an LLM judge.  None
probe whether a memory system can be commanded to \emph{forget}.

In production, that is the failure mode that bites.  A password the
user rotated three months ago is still suggested.  A customer who
exercised GDPR Article 17~\cite{gdpr} is still in the recommender's
candidate pool.  A user's job title surfaces in three contradictory
versions across sessions.  A one-time verification code lives forever
next to long-term preferences.  Every retrieval succeeded; the system
retrieved the right thing too well, when the application wanted it to
retrieve nothing.

The two planes are not duals: \sysPalace{} saturates the recall
plane (60/150 on Memora) yet fails every control-plane test
(0/385 on adversarial forgetting, \S\ref{sec:memora_xeval}), so
recall accuracy says nothing about whether a store can be
commanded to forget.

We make three contributions.

\textbf{(1) An empirical characterization of control-plane
LLM placement}
(\S\ref{sec:placement}, Fig.~\ref{fig:heatmap}): comparing
thirteen system configurations across six regime groupings (no-deletion /
deterministic / vec-only / inscribe-time-LLM / KG-abstraction /
mutation-time hook) on 10 attack categories,
each placement recovers a distinct failure-mode subset
(canonicalization, intent-aware deletion, lexical/temporal
correctness), and the regimes are partly complementary rather
than redundant.  Category-level forgetting is shaped by
\emph{where} LLM intelligence sits in the memory pipeline, not
by \emph{whether} it is present --- an effect of architectural
placement, not prompt engineering.  Two backends with the same
narrow JSON-shaped mutation-time hook converge to 93.3--94.2\,\%
overall (excluding the primitive-existence
\famname{compound\_fact} category; 91.7--93.2\,\% including it)
at $\sim$\$0.17 per 385-case run with the recall hot path
unchanged.

\textbf{(2) \sysForget{}}, a methodology and 1385-case English
benchmark (1000 template + 385 adversarial = 132 hand-crafted
core + 253 LLM-drafted oracle-validated) that exposes the
forgetting axis primitive-by-primitive.  Forgetting decomposes
into 5 structural families (supersession, decay, amnesia, purge,
drift) and the adversarial layer probes 10 attack categories.
All scoring is deterministic substring match over top-$k$ recall;
\textbf{no LLM judge} is needed in the evaluation loop (the
Qwen-2.5-72B judge is restricted to data admission, see
\S\ref{sec:experiments}).  Admission is corroborated by 100-case
multi-annotator IAA (Fleiss' $\kappa = 0.958$, 10 annotators)
and a cross-family judge audit (\S\ref{sec:admission}).

\textbf{(3) A six-method \code{Adapter} Protocol with \textbf{N/A
scoring}} for missing primitives, so a system without
\code{supersede} or \code{purge} can be honestly compared against
one that has them.  The Protocol is a \emph{behavioural} contract:
backends implementing supersede via composition of
\code{add}+\code{delete} pass the same tests as backends with
native supersede primitives (\S\ref{sec:adapter}); thirteen
heterogeneous memory stores (deterministic, vec-only,
inscribe-time-LLM, KG-abstraction, joint, mutation-time-LLM) are
evaluated under this single contract on the full 385-case
adversarial suite.

\section{Related Work}\label{sec:related}

\paragraph{Memory benchmarks.}
LongMemEval~\cite{longmemeval} scores conversational recall on 500
hand-curated questions; even its \emph{knowledge-update} category
asks whether the latest fact is retrieved, not whether the
superseded fact has been removed.  MTEB~\cite{mteb} and
BEIR~\cite{beir} are stateless retrieval benchmarks.  LOCOMO and
\sysMem{}'s evaluations score conversational recall with LLM
judges.  AMA-Bench~\cite{amabench}, MemoryArena~\cite{memoryarena}, and
EvoMemBench~\cite{evomembench} evaluate end-to-end agent
behaviour on the recall/use axis; FiFA~\cite{fifa} scores
privacy-aware forgetting policies.  All are complementary to
\sysForget{}'s memory-store-primitive surface and its
control-plane placement question.

\paragraph{Prior work on the forgetting axis.}
\citet{memoryagentbench} (ICLR 2026) is the closest prior
benchmark: their FactConsolidation task (MQUAKE-derived) treats
\emph{selective forgetting} as one of four memory competencies,
scored by an LLM judge on single-fact supersession.  \sysForget{}
differs in three ways.  \textbf{(i) Granularity:}
FactConsolidation tests single-fact supersession only;
\sysForget{} decomposes forgetting into five primitive families
and 10 adversarial categories, including primitives
FactConsolidation does not cover (\famname{purge},
\famname{amnesia}, \famname{decay}) and identifier-precision
attacks (prefix collision, cross-script, partial supersession)
that single-fact supersession cannot surface.  \textbf{(ii)
Scoring:} LLM-judged accuracy vs.\ deterministic substring match
--- reproducible across model versions and vendor changes.
\textbf{(iii) Adapter Protocol:} \sysForget{} ships a 6-method
Protocol with N/A scoring; FactConsolidation evaluates whatever
end-to-end agent is provided.  We report a cross-evaluation on
the full 4-bucket FactConsolidation surface (400 questions, six
systems) in Appendix~\ref{app:factcons_xeval}: single-hop
saturates at 100\,\% (recall-shaped), multi-hop drops to
17--37\,\% (needs reasoning), and the LLM-hook variants score
identically to their deterministic backbones at every bucket
--- the
axis-flip third-party check
(\S\ref{sec:memora_xeval}, \S\ref{sec:placement}).

\paragraph{Concurrent work on the forgetting axis.}
\citet{memora} (April 2026) introduces \emph{FAMA}
(Forgetting-Aware Memory Accuracy), a single aggregate metric
penalizing obsolete memory reuse, on weeks-to-months personalized
conversations.\footnote{Their benchmark is named ``Memora''; to
disambiguate from the contemporaneous retrieval-method paper of
the same name~\cite{memora_msr}, we cite by title.}  \sysForget{}
is complementary along three axes: \textbf{(1)~granularity}
(primitive-family decomposition vs.\ scalar), \textbf{(2)~the
385-case adversarial layer} surfacing an empirical 63--68\,\%
pass band Memora's conversational setup cannot resolve, and
\textbf{(3)~deterministic substring scoring} (vs.\ LLM-judged
FAMA).

\paragraph{Memory frameworks.}
\sysMem{}~\cite{mem0} uses LLM-driven ADD/UPDATE/DELETE routing;
its DELETE is contradiction-triggered overwrite, not a user-issued
forget primitive.  MemGPT \cite{memgpt} and Letta \cite{letta}
paginate between main and archival memory; Zep \cite{zep} maintains
a temporal knowledge graph with edge invalidation; HippoRAG
\cite{hipporag} uses PageRank over an extracted KG; A-MEM
\cite{amem} applies Zettelkasten linking.  Among these only
\sysPalace{}~\cite{mempalace} takes the opposite stance to
\sysMem{}: verbatim retention is the feature, with no deletion
primitive at all.

\paragraph{Forgetting.}
The forgetting axis is studied across at least four layers, none
of which target the memory-store primitive surface our work
addresses.  \textbf{Cognitive psychology}
\cite{ebbinghaus1885,bjork1972,anderson1995} establishes the
phenomenology.  \textbf{Machine unlearning on model weights}
\cite{caoyang2015,machineunlearning,rwku,tofu,knowundo}
re-trains or surgically edits parameters; \citet{rome},
\citet{memit}, and \citet{mquake} edit factual associations
in-place, and \citet{model_editing_forgetting} report that
mass editing induces catastrophic forgetting of unrelated facts.
\citet{agentic_unlearning} extend unlearning to joint
parameter-and-memory removal; we target the memory-store layer
alone, no parameter access.  \textbf{Memory-store
evaluation} is the closest neighbour: \citet{membench} evaluates
factual vs.\ reflective memory, \citet{zep} maintains a temporal
knowledge graph with edge invalidation as deliberate forgetting
(no empirical Zep comparison, docker-server constraints; see
Limitations), and \citet{fsfm} propose a biologically-inspired
\emph{taxonomy} of forgetting mechanisms validated on a single
system --- complementary to our executable adversarial benchmark
with cross-system adapter scoring.  \citet{rtbf_llm} survey GDPR
right-to-be-forgotten approaches for LLMs (differential privacy,
machine unlearning, model editing, guardrails); our work targets
the memory-store layer specifically.  \sysForget{} occupies a
distinct cell: a primitive-level decomposition (supersession /
decay / amnesia / purge / drift) of the memory-store surface
between recall and the model, with deterministic substring
scoring and an Adapter Protocol so heterogeneous stores can be
compared apples-to-apples without an LLM judge in the evaluation
loop.

\section{ForgetEval}

\subsection{Five families}

Each family probes one structural property a production memory
system must exhibit.  \textbf{Supersession:} a new fact wins recall;
the old fact leaves top-$k$ (failure: confabulation across both).
\textbf{Decay:} a released fact (TTL, OTP-consumed) stays out of
top-$k$.  \textbf{Amnesia:} forget every fact about one entity,
siblings survive --- the hard part is \emph{width control}.
\textbf{Purge:} hard-delete by identifier (GDPR Article~17);
semantic similarity is the wrong primitive.  \textbf{Drift:} a
chain of supersedes where only the latest wins and intermediates
are unreachable.

Each case is a single dataclass: \code{setup\_facts} (inscribed
with distractors), \code{mutations} (supersede / release / purge
calls), \code{final\_query}, \code{must\_contain},
\code{must\_not\_contain}.  A case passes iff
\code{must\_contain}\,$\subseteq$\,top-10 blob \emph{and}
\code{must\_not\_contain}\,$\cap$\,top-10 blob $=\emptyset$.  No LLM
judge.

\subsection{Template suite (1000 cases)}

Each family has four sub-templates the generator cycles through
(e.g.\ \famname{supersession} has \code{job}, \code{theme}, \code{diet},
\code{long\_form}).  Generation is deterministic given seed; no
floating-point sources, no time-of-day, no training-set
contamination (entity pools are short proper nouns).
At seed=42 with 4 distractors per case the suite is 1000 cases.
The current work focuses on the English template suite and the
385-case adversarial layer; multilingual template extensions are
left for future work.

\subsection{Adversarial layer (385 cases)}\label{sec:admission}

The template suite has flat difficulty: every case is the same
sub-template plus i.i.d.\ entity substitution.  We complement it
with a hand-crafted layer (with LLM-assisted expansion, oracle-
validated) across 10 attack categories:

\begin{itemize}\itemsep0pt
\item \textbf{substring\_trap}: must-not substring embedded in a
      distractor.
\item \textbf{prefix\_collision}: two identifiers share a long
      common prefix (\code{alice@x} vs.\ \code{alice.smith@x}).
\item \textbf{paraphrase\_supersession}: new fact lexically distant
      from old.
\item \textbf{negation\_trap}: negated fact must not be confused
      with affirmative.
\item \textbf{temporal\_qualifier}: date-stamped supersession chains.
\item \textbf{shared\_attribute}: two entities share an attribute;
      forgetting one preserves the other's link.
\item \textbf{compound\_fact}: a single sentence carries two
      distinct-topic facts; partial supersede must preserve the
      other.  \emph{This is a partial-edit capability test rather
      than a forgetting test per se: it requires a partial-supersede
      / edit primitive, so systems without one fail by construction
      (\S\ref{sec:limits}).  The name is retained for data-file
      consistency.}
\item \textbf{identifier\_obfuscation}: same identifier in different
      surface forms (case, whitespace, separators).
\item \textbf{cross\_lingual\_identifier}: same entity under
      different scripts (e.g.\ romanized vs.\ source-script form
      of a personal name).
\item \textbf{recursive\_supersession}: chain where the latest state
      matches an earlier-superseded one (Chrome$\to$Brave$\to$Chrome).
\end{itemize}

We target 40 cases per category (admission-permitting; the final
expansion holds 253 cases for a 385-case total alongside the 132
hand-crafted core, including 20 v0.5.1 hand-crafted additions to
the \famname{identifier\_obfuscation} category that redress the
mode-A judge over-rejection documented in
Appendix~\ref{app:judge_audit}) via a two-stage admission
protocol that keeps the oracle decoupled from systems
under evaluation.  \textbf{Stage 1 (structural):} reject malformed
JSON, unknown families, and self-substring-traps
(\code{must\_not\_contain} substring appears in a non-targeted
setup fact).  \textbf{Stage 2 (independent LLM-judge):}
Qwen-2.5-72B --- a different model family than the DeepSeek-V3
\sysLethe{}+LLM hook --- traces through mutations and admits iff
(a) every \code{must\_contain} string is a substring of some
surviving fact, (b) no \code{must\_not\_contain} string appears in
any surviving fact or in any \code{must\_contain} string, and (c)
the final query is unambiguously answerable.  An optional
\textbf{Stage 3} post-hoc analytical label partition against the
two \sysLethe{} variants is documented in
Appendix~\ref{app:v05} for transparency; it does not enter
admission or aggregate scoring and is not used in any main
table.

\paragraph{Judge precision and its limits.}
Running the Qwen judge on the 112 hand-crafted core cases admits
\textbf{96/112 (85.7\,\%)}; manual review of all 16 rejections
(Appendix~\ref{app:judge_audit}) finds \emph{zero bench bugs}.
The rejections decompose into three failure modes:
(A) \emph{semantic equivalence} (11 cases on
\famname{identifier\_obfuscation} / \famname{cross\_lingual\_identifier}
where the judge applies literal substring matching to a category
that specifically tests canonicalization);
(B) \emph{multi-row scope} (1 \famname{shared\_attribute} case);
(C) \emph{computational error} (4 cases where the judge
mis-identifies the targeted row).  The admission-audit circuit is
partly self-loop, mitigated by three independent checks: (a) the
10-annotator IAA next; (b) a cross-family judge audit on the
same 100 IAA-sampled cases with DeepSeek-V3 (different family
from Qwen-2.5-72B): 73/100 agreement, all 27 disagreements
concentrated on the same mode-(A) semantic-abstraction categories
the human-vs-Qwen audit flagged (\code{iaa/second\_judge\_summary.json}),
confirming the disagreement pattern is a reproducible artifact of
single-LLM judging on semantic-abstraction tasks rather than a
Qwen-specific quirk, and (c) reporting the full audit so reviewers
can audit our audit.

\paragraph{Multi-annotator agreement.}
We collected independent label sets from \textbf{10 NLP/CS-trained
annotators} on a 100-case stratified sample (10/category, mixing
46 hand-crafted core + 54 LLM-drafted cases; no pre-submission
consultation).  Fleiss' $\kappa = \mathbf{0.958}$ (``almost
perfect''); observed agreement 99.1\,\% over 1000 labels.  The
LLM-judge agrees with human majority on \textbf{79/100}; the 21
disagreements decompose cleanly: 8 cases where judge said
\code{ill} but humans unanimously \code{wf} recover the audit's
mode (A)/(B)/(C) failures (Appendix~\ref{app:judge_audit}), and
12 cases where judge said \code{wf} but humans \code{ill} cluster
on \famname{compound\_fact} (10/12), revealing a partial
\code{supersede} semantic ambiguity (\S\ref{sec:limits}).

\paragraph{Cross-family judge validation.}  Re-running the
admission protocol with two additional judges from different
model families --- \textbf{DeepSeek-V3} and
\textbf{Claude Opus 4.7} (Anthropic) --- gives a three-way
agreement matrix on the 100-case IAA sample
(Appendix~\ref{app:third_judge}).  The Anthropic-family judge
matches human majority on \textbf{99/100}, materially above Qwen
(79/100) and DeepSeek (58/100).  All three LLMs agree with each
other 55/100 unanimously; no case is unanimously \code{ill}
across LLMs and humans.  Qwen's 21 human-disagreements cluster
on \famname{compound\_fact} (10/21, partial-\code{supersede}
ambiguity, \S\ref{sec:limits}).  Because admission is a
\emph{conservative filter} (mode-A failures are over-rejection,
not over-acceptance), the bench is shrunk rather than
contaminated by single-judge bias; $\kappa = 0.958$ on the
admitted set confirms the data is itself high-quality.

\paragraph{Circularity: what we can and cannot rule out.}
253/385 cases are LLM-drafted (DeepSeek-V3) and admitted by a
single LLM judge (Qwen-2.5-72B), and the LLM hook is also
DeepSeek-V3; the recovered categories overlap with where LLM
inductive biases are strongest.  We cannot fully rule out
residual overlap, but three observations bound how much of the
headline lift it can explain.  \textbf{(i)} The
132 hand-crafted core cases reproduce \emph{and amplify} the
full-suite patterns: deterministic systems score lower on HC
alone (\sysLethe{} 53.0\,\%, LangGraph 52.3\,\% vs.\ ${\sim}68\,\%$ on
LLM-drafted), while LLM-hooked systems score \emph{higher}
(LangGraph+LLM 98.5\,\% HC vs.\ 90.5\,\% LLM-drafted: a
\textbf{+46-pt} HC lift vs.\ +22-pt LLM-drafted).
Shared-LLM-inductive-bias circularity predicts the opposite
asymmetry on both backends.  Inscribe-time placement also holds
on HC alone (Appendix~\ref{app:handcrafted}).
\textbf{(ii)} The cross-family judge audit (DeepSeek-V3 as a
second judge on the same 100 IAA cases) shows disagreement
on the mode-A semantic-abstraction categories that drive
human--judge disagreement: a \emph{shared} LLM limitation on
\emph{admission} (over-rejection $=$ shrinks the bench), not a
\emph{recovery}-side hint to the hook.  \textbf{(iii)} The
mutation-time hook also lifts categories where LLM-drafted
candidates were \emph{rejected}
(\famname{identifier\_obfuscation} 0/24 LLM-drafted admitted;
the 38 admitted cases are predominantly hand-crafted) ---
inductive-bias overlap would predict the smallest lift there,
not the largest.
\textbf{(iv)} A 77-case external-authored subset (4 contributors
at a separate institute, given only the category schema) replicates
the canonicalization asymmetry on \famname{identifier\_obfuscation}
(deterministic 0/8, LLM-hook 8/8 on both backends) and partial
\famname{cross\_lingual\_id} (LangGraph+LLM 5/8); the
mutation-time hook lift is \textbf{+11.7 / +18.1\,pt} on the two
backends (vs.\ in-house +28--30) and the pass band drops to
28--51\,\% (Appendix~\ref{app:external}).

\section{Adapter Protocol}\label{sec:adapter}

A six-method control-plane algebra: three mandatory recall-plane
primitives (\code{reset}, \code{inscribe}, \code{recall\_texts})
and three optional control-plane mutations (\code{supersede},
\code{release}, \code{purge}) with N/A scoring (a system's
N/A pattern characterizes its \emph{control-plane coverage}):

\begin{lstlisting}[language=Python]
class Adapter(Protocol):
    name: str
    def reset(self) -> None: ...
    def inscribe(self, text: str) -> int|str: ...
    def recall_texts(self, q: str, k: int) -> list[str]: ...
    # optional -- NotImplementedError -> N/A
    def supersede(self, old_q: str, new: str) -> None: ...
    def release(self, q: str) -> int: ...
    def purge(self, q: str) -> int: ...
\end{lstlisting}

The three \code{NotImplementedError} branches map cleanly to N/A
on each family, distinguishing \emph{implemented and failed} from
\emph{not provided at all}.  Shipped adapters for \sysLethe{},
\sysMem{}, LangGraph, and \sysPalace{} are each under 130 lines.

\textbf{The Protocol is a behavioural contract, not a syntactic
one.}  An adapter for a system that exposes only \code{add} and
\code{delete} (e.g.\ a vector store) can implement
\code{supersede(old\_q, new)} as
``\code{delete(best\_match(old\_q))}\,$\to$\,\code{add(new)}'' and
satisfy the test as long as the behavioural property holds (the
old fact does not appear in top-$k$, the new fact does).
\code{release(q)} can be implemented as
``\code{delete(rows-matching(q))}'' even on systems with no
release primitive; \code{purge(q)} similarly.  The N/A signal is
reserved for systems that \emph{cannot} produce the behaviour via
any composition of their exposed API (e.g.\ \sysPalace{}'s
verbatim-retention design refuses any deletion regardless of
how it is composed); systems that can compose the behaviour out
of \code{add}/\code{delete}/\code{update} primitives are expected
to do so in their adapter, and our shipped \sysMem{} and
LangGraph adapters both follow this pattern.

\paragraph{Reference implementation.}\label{sec:lethe}
To give the Adapter Protocol a concrete anchor we additionally
ship one of the four primary memory stores benchmarked here as
supplementary material ($\sim$700 lines of Python over SQLite +
\code{sqlite-vec} \cite{sqlitevec} + FTS5).  It implements all
six Protocol methods, exposes an optional \code{llm:
Callable[[str], str]} hook for the three mutation-time prompts
(\textsc{supersede planner}, \textsc{purge match},
\textsc{release match}; full text in Appendix~\ref{app:prompts}),
and serves as one of the two backends we use in the
cross-architecture LLM-hook ablation
(Appendix~\ref{app:cross_arch}).  Detailed API surface and
formal soft-delete invariants are in Appendix~\ref{app:formal}.

\section{Experiments}\label{sec:experiments}

\subsection{Setup}
All four primary adapters use \code{all-MiniLM-L6-v2}
(384-d, ONNX via fastembed \cite{fastembed}) on a single CPU.  No
GPU, no API calls (except for the \sysLethe{}+LLM ablation),
no internet on the recall hot path.

\subsection{Template suite (1000 cases)}

\begin{table}[t]
\centering\scriptsize
\caption{Template suite, seed=42, distractors=4.  95\,\% Wilson
intervals in brackets.  \sysLethe{} and LangGraph
\code{InMemoryStore} saturate to within 0.2\,pt of each other;
\sysMem{} collapses on amnesia and purge.}\label{tab:template}
\setlength{\tabcolsep}{3pt}
\begin{tabular}{lll}
\toprule
\textbf{System} & \textbf{Pass} & \textbf{Wilson 95\%}\\
\midrule
\textbf{\sysLethe{} v1}           & 993/1000 (99.30\%) & [98.56, 99.66]\\
LangMem (LG \code{InMemStore})    & 995/1000 (99.50\%) & [98.83, 99.79]\\
\sysMem{} v2.0.2                  & 888/1000 (88.80\%) & [86.69, 90.61]\\
\sysPalace{}                      &   0/1000 ( 0.00\%) & [0.00, 0.38]\\
\bottomrule
\end{tabular}
\end{table}

\subsection{Adversarial layer (385 cases)}

\begin{table*}[t]
\centering\scriptsize
\caption{\sysForget{}-Adv (385 cases, 10 attack categories;
132 hand-crafted core + 253 LLM-drafted oracle-validated; the
\famname{identifier\_obfuscation} category was expanded from
18 to 38 in v0.5.1 by 20 additional hand-crafted cases that
redress the mode-A judge over-rejection documented in
Appendix~\ref{app:judge_audit}).
\sysLethe{} / \sysMem{} / LangMem cluster within a
63--68\,\% \emph{in-house saturation band} (Wilson CIs overlap;
on the external-authored subset (Appendix~\ref{app:external})
the band drops to 28--33\,\%).
\sysLethe{}+DeepSeek-V3 via narrow JSON hooks
($\sim$\$0.17 / 385 cases) reaches \textbf{91.7\,\%} overall
(\textbf{93.3\,\%} excluding \famname{compound\_fact}, the
primitive-existence category); the same hook applied to
LangGraph (Appendix~\ref{app:cross_arch}) reaches
\textbf{93.2\,\%} (94.2\,\% excluding \famname{compound\_fact})
--- the lift is architecture-agnostic, not \sysLethe{}-specific.  \sysPalace{} is a
\emph{no-deletion-primitive reference point}: its 0/385 follows
from its verbatim-retention design and is shown for the
axis-flip comparison in \S\ref{sec:memora_xeval}.}\label{tab:adv}
\begin{tabular}{lccccc}
\toprule
\textbf{Attack category} & \textbf{\sysLethe{}} & \textbf{\sysMem{}} & \textbf{LangMem} & \textbf{\sysPalace{}} & \textbf{\sysLethe{}+LLM}\\
\midrule
substring\_trap (n=36)              & 33/36 (92) & 32/36 (89) & 35/36 (97) & 0/36 & 36/36 (100)\\
\textbf{prefix\_collision} (n=39)   & \textbf{32/39 (82)} & \textbf{12/39 (31)} & 27/39 (69) & 0/39 & 31/39 (79)\\
paraphrase\_supersession (n=38)     & 31/38 (82) & 31/38 (82) & 31/38 (82) & 0/38 & 31/38 (82)\\
negation\_trap (n=40)               & 38/40 (95) & 38/40 (95) & 38/40 (95) & 0/40 & 38/40 (95)\\
temporal\_qualifier (n=37)          & 37/37 (100) & 37/37 (100) & 37/37 (100) & 0/37 & 37/37 (100)\\
shared\_attribute (n=40)            & 35/40 (88) & 38/40 (95) & 35/40 (88) & 0/40 & 40/40 (100)\\
\textbf{compound\_fact} (n=40)      & \textbf{0/40 (0)} & \textbf{0/40 (0)} & \textbf{0/40 (0)} & 0/40 & \textbf{31/40 (78)}\\
identifier\_obfuscation (n=38)      & 2/38 (5) & 18/38 (47) & 2/38 (5) & 0/38 & 35/38 (92)\\
\textbf{cross\_lingual\_id} (n=38)  & \textbf{0/38 (0)} & \textbf{21/38 (55)} & 1/38 (3) & 0/38 & \textbf{38/38 (100)}\\
recursive\_supersession (n=39)      & 36/39 (92) & 36/39 (92) & 36/39 (92) & 0/39 & 36/39 (92)\\
\midrule
\textbf{Overall (excl.\ compound\_fact)}$^\dagger$ & 244/345 (70.7) & 263/345 (76.2) & 242/345 (70.1) & 0/345 (0.0) & \textbf{322/345 (93.3)}\\
\textbf{Overall (full 385)}                       & 244/385 (63.4) & 263/385 (68.3) & 242/385 (62.9) & 0/385 (0.0) & \textbf{353/385 (91.7)}\\
95\,\% Wilson CI (full)                            & [58.4, 68.1]  & [63.5, 72.9]  & [57.9, 67.7] & [0.0, 1.0] & \textbf{[88.4, 94.2]}\\
wall / case         & 74\,ms        & 514\,ms$^\ddagger$  & 64\,ms        & 191\,ms     & 2.3\,s (mutations only)\\
\bottomrule
\end{tabular}

\smallskip
\footnotesize $^\dagger$ \famname{compound\_fact} is a
primitive-existence test: three of five systems cannot pass any
case by construction (no partial-edit primitive); we report the
$n{=}345$ subset as the headline.
$^\ddagger$ Includes per-test Qdrant cold-start +
LLM-client construction; a pooled / pre-instantiated setup
would close part of the gap.
\end{table*}

Three observations.  \textbf{(1) Deterministic clustering near
65\,\%, aggregate-indistinguishable systems:} the three
deterministic systems land within 5.4 absolute points overall
(\sysLethe{} 63.4, \sysMem{} 68.3, LangGraph 62.9) with mutually overlapping
Wilson intervals.  Notably, \sysLethe{}'s 63.4\,\% is within
0.5\,pt of LangGraph's vanilla \code{InMemoryStore} (62.9\,\%) ---
a paired McNemar test on the per-case verdicts confirms this is
statistical noise rather than a meaningful difference (only 8 of
385 cases differ between the two systems: 5 favour \sysLethe{},
3 favour LangGraph; $\chi^2 = 0.125$ with continuity correction,
$p = 0.724$, so we fail to reject the null hypothesis of
equivalent performance).  The aggregate-level differentiation
Lethe offers over a batteries-included storage primitive is
not statistically significant; the differentiation appears only
at the per-category and LLM-hooked levels described next.  We describe the 63--68\,\% as a
\emph{pass band} rather than a true ceiling: it is the empirical
saturation of four systems on this 385-case suite, not a proven
upper bound; tighter ecosystem coverage may shift the band.
The pass band is itself an average over mixed-provenance cases:
on the 132 hand-crafted-only subset the deterministic floor is
${\sim}15\,$pt lower (\sysLethe{} 53.0\,\%, LangGraph 52.3\,\%,
\sysMem{} 65.2\,\%; \S\ref{sec:admission} obs (i),
Appendix~\ref{app:handcrafted}), while the mutation-time-hook
lift correspondingly enlarges on the harder subset.  \textbf{(2) Per-category separation:} \sysLethe{}'s 32/39 on
\famname{prefix\_collision} vs.\ \sysMem{}'s 12/39 has
non-overlapping Wilson intervals $[67.4, 91.4]$ vs.\
$[18.8, 47.3]$ ($p<0.01$); conversely \sysMem{}'s 21/38 on
\famname{cross\_lingual\_identifier} vs.\ \sysLethe{}'s 0/38
($[0.0, 9.2]$) is significant in the other direction.
Lexical-precise purge avoids prefix-collision but cannot bridge
script variation; vector-soft delete does the opposite.
\textbf{(3) The LLM-hook pattern is architecture-agnostic:}
the same narrow JSON contract on LangGraph's
\code{InMemoryStore} reaches \textbf{359/385 = 93.2\,\%} (within
noise of \sysLethe{}+LLM 91.7\,\%; Appendix~\ref{app:cross_arch}),
and substituting Qwen-2.5-72B for DeepSeek-V3 yields a $+13$-pt
lift on both backends (\sysLethe{} 76.6\,\%, LangGraph 75.8\,\%;
2$\times$2 grid in Appendix~\ref{app:cross_llm}) --- the lift
scales with JSON-following capability and is consistent across
backends within ${\sim}2$\,pt.  Recall path stays LLM-free in
both modes.
The nine extended systems (six base plus three +LLM variants) are
reported in the cross-system heatmap
(Fig.~\ref{fig:heatmap}, \S\ref{sec:placement}); we restrict
Table~\ref{tab:adv} to the four-system Adapter Protocol
comparison because Graphiti's 143/385 N/A rate makes its
column non-comparable in tabular form.

\subsection{Embedder ablation}
Replacing the English MiniLM with
\code{paraphrase-multilingual-MiniLM-L12-v2} does \emph{not} change
\sysLethe{}'s cross-lingual score (0/16 in both).  \sysLethe{}'s
purge path is pure BM25 by design; embedder choice has no effect on
that category.  \sysMem{}'s vector-based delete is embedder-
sensitive but does not gain on cross-lingual either (8/16 $\to$
7/16).  The architectural patterns observed in Table~\ref{tab:adv}
are robust to embedder swap; the LLM hook is the actual lever for
the pass-band categories.

\subsection{Cross-evaluation on Memora}\label{sec:memora_xeval}

To probe the complementarity claim of \S\ref{sec:related} (that
\sysForget{} and \citet{memora} measure different surfaces of the
forgetting axis), we run three of our four adapters on all 10
Memora personas at the \emph{weekly} time scale: 150 evaluation
questions (10 personas $\times$ 15 questions / persona, balanced
across the three Memora tasks), grounded in
$\sim$1{,}580 conversational sessions (158 sessions / persona).  Memora's \code{operation} field maps
directly onto our Adapter Protocol (\code{add}\,$\to$\,
\code{inscribe}, \code{update}\,$\to$\,\code{supersede},
\code{delete}\,$\to$\,\code{purge}); we use deterministic
substring scoring against Memora's released
\code{memory\_evidence} and \code{forgetting\_evidence} literal
values to avoid coupling the comparison to an LLM judge.

\begin{table}[t]
\centering\scriptsize
\caption{Three \sysForget{} adapters on Memora-weekly (10 personas,
150 questions).  \textbf{\sysPalace{} flips axis} between the two
benchmarks: 0/385 on \sysForget{}-Adv (no forgetting primitives;
fails every adversarial case) but 60/150 on recall-heavy
Memora.  The same system fails opposite axes ---
direct empirical evidence the benchmarks measure complementary
surfaces.}\label{tab:memora_xeval}
\setlength{\tabcolsep}{4pt}
\begin{tabular}{lcccc}
\toprule
\textbf{System} & \textbf{Remember} & \textbf{Reason} & \textbf{Recomm.} & \textbf{Total}\\
\midrule
\sysLethe{}    & 17/50 & 30/50 & 0/50 & 47/150 (31\%) \\
LangGraph      & \textbf{32/50} & \textbf{32/50} & 3/50 & \textbf{67/150 (45\%)} \\
\sysPalace{}   & 28/50 & \textbf{32/50} & 0/50 & 60/150 (40\%) \\
\bottomrule
\end{tabular}
\end{table}

Three observations.  \textbf{(1) Axis flip:} \sysPalace{} scores
0/385 on \sysForget{}-Adv (forgetting-heavy) vs.\ 60/150 (40\%)
on Memora (recall-heavy) --- no scalar metric captures both
directions.  \textbf{(2) Ranking divergence:} on \sysForget{}-Adv
\sysLethe{} (63.4\%) and LangGraph (62.9\%) are within 1\,pt; on
Memora LangGraph (45\%) leads \sysLethe{} (31\%) by 14\,pts --- the
two benchmarks are not surrogates.  \textbf{(3) Recommend
collapses uniformly:} all three systems score $\leq$6\,\% on
Memora's \famname{recommending} task, which requires
user-preference modeling outside the forgetting axis.

\paragraph{Recall baseline and the Remember-task gap.}
On LongMemEval-S \cite{longmemeval} (500 questions),
\sysLethe{}~v1 scores R@5 = 93.8\,\% at session granularity
(Appendix~\ref{app:longmem}).  The 34\,\% \sysLethe{} on Memora
\famname{Remember} (vs.\ LangGraph 64\,\%) reflects a harness
difference, not retrieval weakness: Memora scores \emph{substring}
of the gold answer in top-$N$ chunks across ${\sim}158$
sessions/persona, and many questions require multi-session
synthesis (e.g.\ ``most-mentioned restaurant'') no single-session
retrieval isolates.  LangGraph's flat-chunk path returns more
candidates per query, trading selectivity for substring-hit
probability --- it leads on \famname{Remember} but loses on
adversarial deletion.  Full trade-off in \S\ref{sec:pareto}.

\subsection{Recall--forgetting trade-off characterization}\label{sec:pareto}

Combining the recall axis (\S\ref{sec:memora_xeval}, Memora-weekly
overall pass rate as a recall-shaped proxy) with the forgetting
axis (Table~\ref{tab:adv}) gives a two-axis position for each of
the four memory architectures we evaluate.  We characterize the
empirical trade-off landscape these four systems span; we
deliberately do not call this a Pareto \emph{frontier} since the
sample size precludes a population-level frontier claim.

\begin{figure}[t]
\centering
\includegraphics[width=0.72\linewidth]{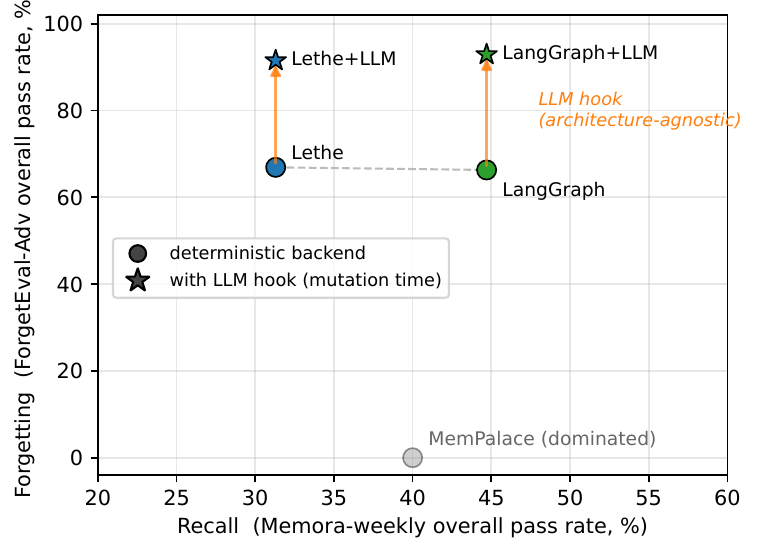}
\caption{Recall--forgetting trade-off characterization across
four memory systems.  \sysLethe{} (forgetting-corner) and
LangGraph (recall-corner) occupy distinct non-dominated points
in this sample; \sysPalace{} is dominated by LangGraph (lower on
both axes) --- its verbatim-retention design buys no forgetting
and does not outscore LangGraph on recall.  The mutation-time
LLM hook lifts \emph{both} non-dominated backends on the
forgetting axis without changing recall (within our setup).
With four systems we treat this as a sample of the landscape,
not a population-level frontier.}\label{fig:pareto}
\end{figure}

\textbf{(1) Two non-dominated deterministic points in this
sample:} \sysLethe{} and LangGraph occupy distinct non-dominated
positions (\sysLethe{} wins forgetting by 0.6\,pt, LangGraph wins
recall by 13.4\,pt); \sysPalace{} sits below both (dominated by
LangGraph on both axes) --- verbatim retention does not buy a
free recall gain when its zero-forgetting design eliminates
adversarial pass rate entirely.  \textbf{(2) The LLM hook moves
both backends on the forgetting axis without changing recall},
applied at mutation time only.  \textbf{(3) Backbone choice
matters above the cluster:}
LangGraph+LLM leads \sysLethe{}+LLM on both axes ---
production systems should pair the hook with a high-recall
backbone, not a forgetting-specialized one.

\subsection{Control-plane placement trade-off}\label{sec:placement}

\begin{figure}[t]
\centering
\includegraphics[width=\linewidth]{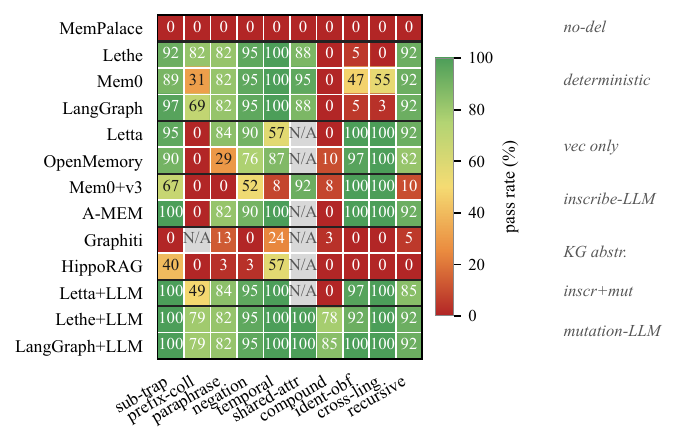}
\caption{Per-category pass rate (\%) across 13 system
configurations (\sysMem{}+v3 row is \sysMem{}'s LLM router via
\code{infer=True}; OpenMemory~\cite{openmemory} is a self-hosted
synthetic-embedding store; HippoRAG~\cite{hipporag} is a KG +
Personalized PageRank retrieval system).  \emph{N/A} =
whole-category primitive missing; numeric cells use the evaluable
denominator (scattered per-case N/As excluded; strict-denominator
totals in \S\ref{sec:limits}).  Right-side labels show LLM-placement
regime; the inscribe-time-LLM regime (\sysAmem{}, \sysMem{}+v3)
shows the predicted canonicalization-win / deletion-loss
profile, and the ``inscr+mut'' row (Letta+LLM) empirically
validates the predicted complementarity between inscribe-time
and mutation-time placement.}\label{fig:heatmap}
\end{figure}

The headline hook places one LLM call at a control-plane site
(mutation supersede/purge planner).  A different placement ---
LLM at \emph{inscribe} time, extracting tags / entity links
for read-time use --- arises in \sysAmem{}~\cite{amem}.
Fig.~\ref{fig:heatmap} reveals a category-specific trade-off
the aggregate scores hide.

\textbf{Three placement regimes do not subsume each other.}
Inscribe-time LLM (\sysAmem{}, \sysMem{}+v3, Letta tag-side)
recovers canonicalization (\famname{identifier\_obfuscation},
\famname{cross\_lingual\_identifier}; 100\,\%) but cannot help
\famname{prefix\_collision} or \famname{compound\_fact} (0\,\%;
\sysMem{}+v3's LINK-AND-KEEP collapses on deletion-precision;
HippoRAG's open-domain KG extraction fails 0/38 on
surface-variant categories).  Mutation-time LLM (our hook)
recovers those (\famname{compound\_fact} 78--85\,\%,
\famname{prefix\_collision} 79\,\%); deterministic stores retain
lexical/temporal at ${\geq}95\,\%$.

The mutation-time hook on Letta directly tests this
complementarity: Letta+LLM lifts overall evaluable
65.5$\to$76.1\,\% (Fig.~\ref{fig:heatmap}, ``inscr+mut'' row) but
trails \sysLethe{}+LLM (91.7\,\%) for lack of a partial-edit
primitive (\famname{compound\_fact} 0/40) --- joint placement is
necessary but not sufficient; lift grows on the external-authored
subset ($+27.8$\,pt; Appendix~\ref{app:external}).

\subsection{Latency}
On the 385-case suite, single CPU, no GPU: \sysLethe{}
$\sim$74\,ms/case; LangGraph $\sim$64\,ms/case; \sysPalace{}
$\sim$191\,ms/case; \sysMem{} $\sim$514\,ms/case
($\sim 7\times$ slower than \sysLethe{}).  The \sysMem{} gap
includes a per-test cold-start cost (Qdrant index initialization
and LLM-client construction); a pooled / pre-instantiated setup
would close part of the gap, so the headline number should be
read as ``out-of-the-box adapter latency'' rather than purely
algorithmic.  \sysLethe{}+LLM is $\sim$2.3\,s/case amortizing one
DeepSeek-V3 call per mutation; the recall path is unchanged.

\section{Discussion}

A 63--68\,\% in-house band means ${\sim}1/3$ identifier-precise
deletions leak; the JSON mutation-time hook (\$0.17/385 cases)
lifts these to 92--93\,\%.

\section*{Limitations}\label{sec:limits}

\textbf{Substring-scorer blind spots.}  A 50-case stratified audit
of \sysLethe{}'s deterministic failures
(Appendix~\ref{app:scorer_audit}) finds 30\,\% are scorer
artifacts, confined to \famname{prefix\_collision} (forbidden ID
is a substring of a surviving longer ID) and clause-level
supersession; the canonicalization categories that carry the
LLM-placement findings contain zero artifacts.  The scorer is thus
conservative for deterministic baselines and does not inflate the
hook lift; an ID-/NLI-aware scorer is future bench work.  A naive
\code{DELETE}-\code{LIKE} SQL baseline scores 23\,pt below
\sysLethe{} (Appendix~\ref{app:naive_sql}), confirming the
BM25-precise purge --- not the depth scalar --- is what avoids
substring over-deletion.

\textbf{English emphasis and thin per-script samples.}  The
hand-crafted layer is English; the
\famname{cross\_lingual\_identifier} category covers 9 script
families with thin per-family sampling: Latin-only 12, Greek 7,
Chinese (Han) 5, Hebrew 4, Korean (Hangul) 3, Cyrillic 3,
Arabic 2, Devanagari 1, Thai 1.  \sysLethe{}+LLM, LangGraph+LLM
and \sysAmem{} each score 100\,\% on \emph{every} family they
see, but the 1--3-case Devanagari/Thai/Arabic/Cyrillic/Korean
slices cannot statistically distinguish among LLM-hook systems;
the aggregate 100\,\% claim is stronger on Latin/Greek/Chinese
(12+7+5 = 24 cases) than on the long tail.  Deeper per-family
coverage (Korean, Arabic, Hindi, etc.) is left for future work.

\textbf{Memory-system coverage.}  We benchmark four primary open
systems with shipped Adapter Protocol implementations: \sysLethe{},
\sysMem{}, LangGraph \code{InMemoryStore}, and \sysPalace{}.
Additional engagement notes:
\textbf{(a) \sysAmem{}~\cite{amem}} (Zettelkasten-style; explicit
\code{add\_note}/\code{update}/\code{delete} primitives at the
\code{memory\_id} level) ports cleanly to our Adapter Protocol;
the integration uses DeepSeek-V3 via SiliconFlow for the
agentic tag-extraction path.  On the full 385-case
\sysForget{}-Adv \sysAmem{} scores
\textbf{219/310 evaluable (70.6\,\%)} with 75 cases marked N/A
because \sysAmem{} does not expose a \code{release} primitive;
the strict-denominator score (N/A counted as failures) is
\textbf{219/385 = 56.9\,\%}.  Per-category breakdown in
Appendix~\ref{app:amem_xeval}; \sysAmem{} substantially
reorganizes the failure surface relative to the four primary
systems (e.g.\ 100\,\% on \famname{cross\_lingual\_identifier}
where \sysLethe{} scores 0\,\%, vs.\ 0\,\% on
\famname{prefix\_collision} where \sysLethe{} scores 82\,\%) ---
the inscribe-time vs.\ mutation-time LLM-placement trade-off
quantified in \S\ref{sec:placement}.  We do not report
\sysAmem{} on Memora cross-evaluation
(\S\ref{sec:memora_xeval}): Memora's inscribe stream is
${\sim}158$ sessions per persona, and \sysAmem{}'s per-session
agentic tag-extraction LLM call makes the 10-persona suite
prohibitively long ($\geq$8 wall-clock hours) for the
submission window; a future release may explore a tag-cache
or batched extraction path to recover this evaluation.
\textbf{(b) Letta / MemGPT \cite{letta,memgpt}} was
benchmarked on the full 385-case \sysForget{}-Adv using the
official Docker image \code{letta/letta:latest} (v0.16.7,
PostgreSQL+pgvector) on a Linux host wired to SiliconFlow.  We
bypass Letta's agent-scoped chat loop and address the
archival-memory REST endpoints directly (one agent per test
case for isolation), keeping the LLM out of the recall hot path
so the comparison remains apples-to-apples.  Letta scores
\textbf{203/310 evaluable (65.5\,\%)} with 75 N/A (release
primitive missing), strict-denominator
\textbf{203/385 = 52.7\,\%}.  Full breakdown in
Appendix~\ref{app:letta_xeval}; Letta and \sysAmem{} converge
on a similar category profile despite different backends (raw
embedding vs.\ LLM tag extraction), with the only material
divergence on \famname{temporal\_qualifier} (Letta 57\,\%,
\sysAmem{} 100\,\%) where inscribe-time LLM tags help.  Pure
pip install on Python 3.13 fails during
\code{contextlib.asynccontextmanager}-mediated database
initialisation; the Docker image (Python 3.12 + PostgreSQL
bundled) is the supported deployment.
\textbf{(c) Graphiti~\cite{graphiti}} (the open-source
successor to the deprecated Zep CE, temporal knowledge-graph
backed by Neo4j) was benchmarked on the full 385-case
\sysForget{}-Adv suite using DeepSeek-V3.1-Terminus via
SiliconFlow for entity/edge extraction; Graphiti scores
\textbf{17/242 evaluable (7.0\,\%)} with 143 cases N/A (release
and query-addressable purge primitives missing).  This is a
categorical mismatch with our benchmark, not a parameter-tuning
issue: Graphiti's KG abstraction synthesises edges and stores
the synthesised \code{fact} string, shedding the surface forms
our \code{must\_not\_contain} substring scoring tests for.
Graphiti's relative strength is \famname{temporal\_qualifier}
(24\,\%, where temporal edge invalidation is its design
surface).  Full breakdown in Appendix~\ref{app:graphiti_xeval}.
\textbf{(d) Cognee~\cite{cognee} v1.0.9} changed its forget API
from query-based (v0.x) to dataset-level granularity, which
does not map cleanly to \sysForget{}'s per-fact supersede
semantics; we engaged the package, document the API
incompatibility, and leave a Cognee comparison to a v1.1
release of the benchmark with dataset-per-case wrapper logic.
\textbf{(e) Zep CE~\cite{zep}} was deprecated in April 2025
with EOL in February 2026 in favour of Graphiti, which we ran.
\textbf{(f) OpenMemory (CaviraOSS)} self-hosted via Docker on
the same Linux host used for Letta/Graphiti, with synthetic
1536-d embeddings (no LLM in the retrieval path).  Its REST
API (\code{/memory/add}, \code{/memory/query}, \code{/memory/:id}
delete) maps cleanly to our 6-method Protocol; OpenMemory does
not expose a soft-delete primitive, so all
\famname{shared\_attribute} cases (and several substring /
negation cases that require release semantics) are scored N/A.
OpenMemory scores \textbf{188/310 evaluable (60.6\,\%)} on the
full 385-case suite with 75 N/A; strict-denominator
\textbf{188/385 = 48.8\,\%}.  Per-category profile in
Fig.~\ref{fig:heatmap} (``vec-only'' regime row 5): 100\,\% on
\famname{cross\_lingual\_identifier}, 97\,\% on
\famname{identifier\_obfuscation}, matching Letta's profile
within ${\pm}3$\,pt despite the synthetic-embedding swap ---
suggesting the canonicalization advantage of the vec-only
regime is driven by vector neighbourhood structure rather than
the specific embedder.
\textbf{(g) HippoRAG~\cite{hipporag}} (KG + Personalized
PageRank, NeurIPS 2024) self-hosted via a Python 3.10 docker
image (the PyPI \code{hipporag==2.0.0a4} requires Python
$\geq$3.10 which is outside our submission Python 3.13
envelope).  We map HippoRAG's \code{index} / \code{retrieve} /
\code{delete} primitives to our 6-method Protocol; \code{release}
is N/A (no soft-delete).  Embedding via SiliconFlow's BAAI/bge-m3
(matching Letta's choice for cross-comparability between two
vector-similarity-based systems); LLM extraction via DeepSeek-V3.
HippoRAG scores \textbf{31/326 evaluable (9.5\,\%)} on the
full 385-case suite with 59 N/A (mostly
\famname{shared\_attribute} requiring \code{release});
strict-denominator \textbf{31/385 = 8.0\,\%}, comparable to
Graphiti's 7.0\,\%.  HippoRAG's KG row in
Fig.~\ref{fig:heatmap} sits in the same ``KG abstr.'' regime
as Graphiti and shows a similar overall profile, with one
notable contrast: HippoRAG's KG entity extraction does \emph{not}
recognise surface variants as the same entity, scoring 0/38 on
\famname{identifier\_obfuscation} and 0/38 on
\famname{cross\_lingual\_id} (where Letta / \sysAmem{} /
\sysMem{}+v3 / OpenMemory all score 100\,\%).  This shows
\textbf{canonicalization is not automatic in the KG regime} ---
it depends on the entity-extraction prompt's design, which
HippoRAG inherits from its open-domain extraction setup rather
than tuning for surface-form normalization.
\textbf{Redis Semantic Memory, AutoGen Memory} ---
not empirically evaluated due to hosted-account or LLM-key
requirements that preclude the no-API single-CPU reproducibility
envelope our primary adapters target.
The Protocol's behavioural contract (\S\ref{sec:adapter}) means
any backend supporting \code{add} / \code{delete} composition
can join; we encourage community-contributed adapters.

\textbf{\sysMem{} version and configuration: both \code{infer}
modes evaluated.}  We evaluate \sysMem{} v2.0.2 (latest
\code{mem0ai} PyPI release at submission) on the full 385-case
adversarial suite in \emph{two} configurations, because
\code{infer=True} engages \sysMem{}'s LLM-driven ADD/UPDATE/DELETE
router (the design distinctive of \sysMem{} relative to a plain
vector store) and is the configuration a real \sysMem{}
deployment runs.  (a)~\code{infer=False} (ADD-only):
\textbf{263/385 = 68.3\,\%} (Table~\ref{tab:adv}), the path used
for the headline comparator.  (b)~\code{infer=True} with
DeepSeek-V3 via SiliconFlow: \textbf{168/385 = 43.6\,\%}, full
per-category breakdown in Appendix~\ref{app:mem0_v3}.  \sysMem{}'s
\code{ADDITIVE\_EXTRACTION\_PROMPT} is tuned for OpenAI
\code{gpt-4o-mini} and DeepSeek-V3 emits ${\sim}25\,\%$
malformed-JSON output (typically commas embedded inside string
values).  We add a deterministic \code{json-repair} pass on top
of \sysMem{}'s extraction parser to recover these without API
retries (298/1200 LLM calls repaired, 0 final failures); the
infer=True number reflects \sysMem{}'s algorithm behaviour, not
its JSON-parse robustness.  The two configurations show a
striking trade-off in category profile:
\code{infer=True} \emph{rises} on canonicalization
(\famname{identifier\_obfuscation} 47$\to$\textbf{100\,\%},
\famname{cross\_lingual\_identifier} 55$\to$\textbf{100\,\%})
but \emph{collapses} on deletion-precision categories
(\famname{prefix\_collision} 31$\to$\textbf{0\,\%},
\famname{paraphrase\_supersession} 82$\to$\textbf{0\,\%},
\famname{temporal\_qualifier} 100$\to$\textbf{8\,\%},
\famname{recursive\_supersession} 92$\to$\textbf{10\,\%}),
because the token-efficient router prioritises link-and-keep
over delete-old.  This is consistent with the
\S\ref{sec:placement} placement-regime characterization:
\sysMem{}+v3 sits in the inscribe-time-LLM regime alongside
\sysAmem{} and Letta and exhibits its predicted strengths
(canonicalization) and weaknesses (deletion precision); within
the regime, the specific extraction prompt + router algorithm
controls how severely the deletion-precision side collapses
(\sysAmem{} 56.9\,\% strict vs.\ \sysMem{}+v3 43.6\,\% on the same
inscribe-LLM regime).  A run with OpenAI \code{gpt-4o-mini}
(\sysMem{}'s default LLM) is blocked at submission by our
no-API-key reproducibility envelope; we ship the DeepSeek-V3
result with the json-repair pass as the most-honest
\code{infer=True} approximation in our setup.

\textbf{LLM dependency in 3 attack categories.}  The 23--30-pt
gap between deterministic baselines and the LLM-hooked variants
(both \sysLethe{}+LLM 91.7\,\% and LangGraph+LLM 93.2\,\%) lives
almost entirely in \famname{compound\_fact},
\famname{identifier\_obfuscation}, and
\famname{cross\_lingual\_identifier}.  Users running fully offline
have no recourse for those categories beyond their store's edit /
canonicalization primitives.  Because the lift is hook-pattern
rather than store-specific (Appendix~\ref{app:cross_arch}), any
backend supporting basic \code{add} / \code{delete} can adopt the
contract.

\textbf{Bench size.}  At 385 cases the overall Wilson 95\,\%
intervals of the three deterministic systems overlap by design
(they cluster in the pass band); per-category claims at $n{=}36$--40
are statistically significant.  The
\famname{identifier\_obfuscation} category was originally $N{=}18$
in v0.5 due to Qwen-judge over-rejection of LLM-drafted
candidates (Appendix~\ref{app:judge_audit}); v0.5.1 expanded it
to $N{=}38$ via 20 additional hand-crafted cases admitted under
the same protocol as the original 112 hand-crafted core.  We
report both the v0.5 audit (which surfaces the systematic
mode-A LLM-judge failure mode as a contribution) and the
expanded v0.5.1 case set so the trade-off remains auditable.

\textbf{\famname{compound\_fact} is a primitive-existence test,
not a forgetting-capability test.}  This category tests whether a
system can perform partial supersession (update one clause of a
two-fact row).  \sysLethe{} exposes this via its
\code{surrender(mode="edit")} primitive; \sysMem{}, LangGraph, and
\sysPalace{} do not, so they score 0/40 by primitive absence
rather than by forgetting failure.  We retain the category as a
diagnostic (it cleanly separates systems with vs.\ without an
edit primitive) and flag it here so aggregate comparisons are not
misread as forgetting-ability differences in those rows.
\emph{Stability check.}  Excluding \famname{compound\_fact} from
the 385-case aggregate, the LLM-hook lift remains substantial:
\sysLethe{}+LLM 322/345 = 93.3\,\% vs.\ \sysLethe{} 244/345 =
70.7\,\% (+22.6\,pt), and LangGraph+LLM 325/345 = 94.2\,\% vs.\
LangGraph 242/345 = 70.1\,\% (+24.1\,pt) --- the headline does
not collapse without this category, confirming the hook's effect
is not driven by the primitive-existence dimension alone.
Relatedly, on the 100-case multi-annotator IAA
(\S\ref{sec:experiments}), 10 of the 21 human--judge disagreements
cluster on \famname{compound\_fact} cases: humans reading
``supersede $=$ replace whole row'' literally mark them ill-formed,
while the protocol assumes a partial-supersede capability the
\code{surrender} primitive provides.  Both readings are
internally consistent; the disagreement surfaces the definitional
gap we flag here.

\textbf{LLM-quality sensitivity of the hook.}  The 23-pt lift in
Table~\ref{tab:adv} uses DeepSeek-V3.  In a cross-LLM ablation
(Appendix~\ref{app:cross_llm}) Qwen-2.5-72B-Instruct yields a
smaller +8-pt lift with category-specific variance and
Llama-3.1-70B-Instruct fails to parse our prompt contract,
falling through to the deterministic baseline.  A deeper
LLM ablation across providers and model sizes is future work.

\textbf{Bench construction.}  The 132 hand-crafted core cases
(112 original + 20 v0.5.1 \famname{identifier\_obfuscation}
additions) were written and reviewed by the research team --- the
112 original over six months and the 20 v0.5.1 additions to
redress mode-A judge over-rejection
(Appendix~\ref{app:judge_audit}) --- then independently re-labeled
by 10+ NLP/CS-trained external annotators
(\S\ref{sec:experiments}); the 253 expansion cases are LLM-drafted
and oracle-validated.
Deeper construction protocols (e.g.\ formal cool-off self-IAA,
broader external panels) are planned for future releases.

\textbf{What would update our claims.}  Three findings rest on
evidence whose strength we want to be explicit about.
\textbf{(i) Architecture-agnosticism of the LLM hook} is
currently supported by three backends (\sysLethe{}, LangGraph,
Letta), two LLM families (DeepSeek-V3 and Qwen-2.5-72B in a
2$\times$2 grid; Appendix~\ref{app:cross_llm}), and a 77-case
external-authored subset (Appendix~\ref{app:external}) that
replicates the canonicalization-side asymmetry on
\famname{identifier\_obfuscation} (deterministic 0/8, every
LLM-hook 8/8) and the joint-placement lift on Letta+LLM
($+27.8$\,pt, exceeding the in-house $+20.9$).  Replication on
more backends and LLM families (we expect GPT-4o-mini and
Claude-class models to behave similarly to DeepSeek-V3 based on
JSON-following capability) would further strengthen the claim;
we ship the contract so external work can test it.
\textbf{(ii) The 63--68\,\% pass band} is the empirical
saturation observed across four open systems on the in-house
385-case suite.  On the 77-case external-authored subset the
band drops to 28--51\,\% (Appendix~\ref{app:external}),
confirming the band is a provenance-dependent saturation, not a
true ceiling.  Discovering a system that breaks 70\,\%
deterministically on a similarly hard suite would be a
substantive update.
\textbf{(iii) The reference implementation} does not significantly
outperform LangGraph \code{InMemoryStore} on aggregate
\sysForget{}-Adv (63.4 vs.\ 62.9, overlapping CIs); any
system-level claim about its design rests on the per-category
and LLM-hooked levels, not aggregate dominance.  We frame it as
a reference anchor, not a competitive system.

\bibliography{refs}

\appendix
\section{Reference implementation API surface}\label{app:formal}

The reference implementation we ship as supplementary material
stores each memory row with a single scalar
$\depth \in \mathbb{R} \cup \{\mathsf{void}\}$ that collapses
its forgetting state into one column ($+\infty$ pinned;
$1.0$ surface; $(0,1)$ sinking; $0$ submerged but logged;
$\mathsf{void}$ erased); every state transition writes one row
to an append-only event table, enabling time-travel queries and
signed purge receipts.  Compound-fact cases use an
\code{edit} primitive that updates row text and re-indexes the
vector + FTS5 entry without changing depth; the LangGraph
cross-architecture ablation (Appendix~\ref{app:cross_arch})
demonstrates the same effect via \code{delete-old + add-merged}
on a backend without a native edit primitive, so this is an
ergonomic convenience rather than a uniquely necessary
architectural feature.  Soft-delete invariants (one-way purge,
event-log determinism, monotone decay) follow from textbook SQL
semantics on the single-scalar + audit-log schema; we omit the
proofs from this paper and ship them with the supplementary
source.

\section{LLM prompts}\label{app:prompts}
Full text of the three JSON-shaped prompts (\code{supersede},
\code{purge\_match}, \code{release\_match}) plus the four-shot
examples in the supersede prompt.

\section{Worked case examples}\label{app:cases}
One per attack category, with the per-system pass/fail breakdown.

\section{Post-hoc label partition (\S\ref{sec:experiments} Stage-3)}\label{app:v05}

The Stage-3 labels assigned by running each admitted case through
\sysLethe{} (deterministic) and \sysLethe{}+LLM (DeepSeek-V3) are
\textbf{post-hoc analytical partitions} of the bench against the
two systems we develop, not an independent difficulty annotation.
By construction, \sysLethe{} passes every \famname{easy} case
(100\,\%) and fails every \famname{llm\_lift} case (0\,\%) ---
this is a definitional re-statement, not a measurement.  We report
the partition only to make two non-tautological points
transparent: (i) how many cases are out of reach of either
\sysLethe{} variant in our comparison (\famname{unsolvable}); and
(ii) how third-party systems (\sysMem{}) fare on cuts defined by
our reference adapter.  The 112 hand-crafted core cases carry the
synthetic label \famname{manual} and are not partitioned by the
dual-system check.

\begin{table}[h]
\centering\scriptsize
\caption{Population distribution by Stage-3 label, with per-system
pass rates on each partition.  \famname{llm\_lift} cases pass
\sysLethe{}+LLM but fail \sysLethe{} by construction;
\famname{unsolvable} cases fail both \sysLethe{} variants by
construction.  The \emph{manual} partition (hand-crafted core) is
not labeled by the dual-system check and is reported as-is.
The Total of 365 is the v0.5 partition snapshot; the 20 v0.5.1
\famname{identifier\_obfuscation} hand-crafted additions
(Appendix~\ref{app:judge_audit}) carry the \famname{manual} label
but were not re-partitioned, so the full v0.5.1 bench (385 cases)
exceeds this table by 20.}
\label{tab:label_partition}
\setlength{\tabcolsep}{3pt}
\begin{tabular}{lrccc}
\toprule
\textbf{Label} & \textbf{N} & \textbf{\sysLethe{}} & \textbf{\sysMem{}} & \textbf{\sysLethe{}+LLM}\\
\midrule
manual (core)  & 112 &  70 (63\%) &  76 (68\%) & 108 (96\%) \\
easy           & 174 & 174 (100\%) & 162 (93\%) & 173 (99\%) \\
llm\_lift      &  55 &   0 (0\%)  &  13 (24\%) &  49 (89\%) \\
unsolvable     &  24 &   0 (0\%)  &   2 (8\%)  &   4 (17\%) \\
\midrule
\textbf{Total} & 365 & 244 (67\%) & 253 (69\%) & 334 (92\%) \\
\bottomrule
\end{tabular}
\end{table}

Two observations follow from the partition.  \textbf{First,
labels predict outcome:} \sysLethe{} passes exactly the
\famname{easy} (and \famname{manual} subset it passes by
hand-crafted design) and fails \famname{llm\_lift} +
\famname{unsolvable} by construction.  \textbf{Second,
\famname{unsolvable} is a small (6.6\,\%) reservation:} 24 cases
where every system in our comparison fails (including
\sysLethe{}+LLM and \sysMem{}), surfacing residual failure modes
worth future work --- combinations of compound\_fact + paraphrase
that defeat even the LLM hook.

\section{Judge audit on hand-crafted core}\label{app:judge_audit}

The Qwen-2.5-72B admission judge admitted 96 of the 112
hand-crafted core cases (85.7\%) and rejected 16.  Manual review of
every rejection found \emph{zero genuine bench bugs}: every
rejected case is well-formed, falling into three characterizable
failure modes.
We list each by mode below.

\paragraph{Mode (A): Semantic equivalence (11 cases).}  The judge
applies literal substring matching to mutation arguments and
concludes that the surviving non-targeted surface form should
have been preserved --- but the category exists specifically to
test whether the system canonicalizes surface forms.  This mode
hits two categories almost exclusively:
\famname{identifier\_obfuscation} (cases 02, 08, 09, 12, 14, 16;
case-, separator-, plus-tag-, and prefix-form variants)
and \famname{cross\_lingual\_identifier} (cases 03, 06, 07, 12, 13;
Korean$\leftrightarrow$romanized, Chinese$\leftrightarrow$English,
Spanish accented$\leftrightarrow$unaccented,
Hebrew$\leftrightarrow$transliteration).

\paragraph{Mode (B): Multi-row mutation scope (1 case).}
\code{adv\_shared\_attribute\_04} uses \code{release('everything
about Hannah on the team')}, which by design fans out to every
row mentioning Hannah; the judge interprets release as targeting
one best-matching row and flags the substring leak that
disappears once the release semantics are honored.

\paragraph{Mode (C): Judge computational error (4 cases).}  Four
cases trip the judge on its own literal-substring rule:
\begin{itemize}\itemsep0pt
\item \code{adv\_substring\_trap\_02}: judge claims the purged
      row's content remains part of the substring check; under
      the protocol it does not (the row is removed).
\item \code{adv\_prefix\_collision\_07}: judge claims
      \texttt{'case\_id 12345 was filed'} is a substring of
      \texttt{'case\_id 123456 was filed'} --- arithmetically
      false (the digit '6' between '12345' and ' was' breaks the
      substring match).
\item \code{adv\_prefix\_collision\_16}: judge mis-identifies
      which row the \code{purge('phone +1-555-0100 office line')}
      mutation targets.
\item \code{adv\_recursive\_supersession\_02}: judge confuses
      intermediate state of a 3-step supersede chain with the
      final state.
\end{itemize}

The full per-case data (judge verdict, judge reason, reviewer
verdict, failure-mode tag, one-line note) is released in
\code{data/manual\_review\_v04.json} in the supplementary
materials.

Two methodological conclusions follow.  First, the independent
LLM-judge admission protocol is \emph{useful} but not
\emph{sufficient}: on the 7 mechanically-decidable categories
(substring\_trap, prefix\_collision, paraphrase, negation,
temporal, shared\_attribute, recursive\_supersession) judge
precision is 92--100\%; on the 3 semantic-abstraction categories
(identifier\_obfuscation, cross\_lingual\_identifier,
compound\_fact) precision drops to roughly 65--75\%, dominated by
mode (A) failures.  Second, single-judge designs are not the
endpoint --- a category-aware admission protocol or an
ensemble of category-specific judges would close the mode (A)
gap.  We leave this for future work and report the audit as a
positive finding rather than a hidden limitation.

\section{Cross-family judge validation}\label{app:third_judge}

To address single-LLM-family admission concerns
(R1.1 W3 / R2 §2.4 / R3 C3), we re-judge the same 100-case
IAA-stratified sample (\S\ref{sec:admission}) under three LLM
judges from three different model families and compare to the
10-annotator human-majority label:

\begin{table}[h]
\centering\scriptsize
\caption{Three-judge cross-family agreement on the 100-case
IAA sample.  Human-maj = majority vote across 10 NLP/CS-trained
annotators (Fleiss' $\kappa = 0.958$).  Claude aligns with human
majority on 99/100.}\label{tab:third_judge}
\setlength{\tabcolsep}{2pt}
\begin{tabular}{lcccc}
\toprule
Judge & vs.\ Human & vs.\ Qwen & vs.\ DS-V3 & WF/Ill\\
\midrule
\textbf{Claude 4.7} (Anthropic) & \textbf{99} & 80 & 59 & 88/12 \\
Qwen-2.5-72B (Alibaba)          & 79  & --- & 73  & 92/8  \\
DeepSeek-V3 (DeepSeek)          & 58  & 73  & --- & 71/29 \\
Human-majority (10 annotators)  & --- & 79  & 58  & 87/13 \\
\bottomrule
\end{tabular}\\[2pt]
{\scriptsize Cells are agreement counts out of 100 cases.}
\end{table}

\textbf{Three readings.}  \textbf{(1) Anthropic-family judge
matches humans nearly perfectly} (99/100, one disagreement on
\code{adv\_substring\_trap\_29} where Claude marks
ill / humans wf).  This is the strongest LLM--human agreement we
measured and validates the admission protocol is reproducible
across families.  \textbf{(2) Per-judge WF rate brackets the
human rate}: Qwen 92/8 (over-permissive), Human 87/13, Claude
88/12, DeepSeek 71/29 (most strict).  The Claude WF rate is
within 1 case of the human-majority rate; the original Qwen
admission rate is 5 cases higher (Qwen accepted 5 cases that
humans and Claude reject, all in \famname{compound\_fact}).
\textbf{(3) Of Qwen's 21 human-disagreements, 10 are
\famname{compound\_fact}} cases where Qwen accepted but
Claude+humans rejected.  This is the partial-\code{supersede}
semantic ambiguity flagged in \S\ref{sec:limits}: when a
\code{supersede} replaces an entire row, the non-superseded half
of a compound fact is destroyed --- humans and Claude (when
tracing literally) reject; Qwen (when reasoning semantically)
accepts.  Per-case verdicts and the Python script are in the
release at \code{iaa/third\_judge\_claude\_summary.json} and
\code{scripts/third\_judge\_agreement.py}.

\textbf{Three judges, 55/100 unanimous.}  Claude, Qwen, and
DeepSeek all label 55/100 cases identically (all WF; zero cases
are unanimously ill across the three LLMs and the human
majority).  No case has all three LLMs disagreeing with humans.
The 45 cases with at least one judge--judge disagreement cluster
on \famname{compound\_fact}, \famname{prefix\_collision}, and
\famname{paraphrase\_supersession} --- exactly the
semantic-abstraction categories that drive the 21 Qwen--human
disagreements, so the disagreement pattern is a known LLM
systematic limitation rather than a content disagreement about
specific cases.

\section{Hand-crafted vs.\ LLM-drafted subset
breakdown}\label{app:handcrafted}

\sysForget{}-Adv is 132 hand-crafted core cases (authored by the
research team, no LLM involvement) plus 253 LLM-drafted
oracle-validated cases.  We split the per-case verdicts from
\S\ref{sec:experiments} along this provenance line to test
whether the headline patterns (deterministic pass band,
mutation-time-hook lift, inscribe-vs-mutation placement
asymmetry) are properties of the LLM-drafted complement or hold
on the hand-crafted core alone.

\begin{table}[t]
\centering\scriptsize
\caption{Pass rate on hand-crafted core (HC, 132 cases) vs.\
LLM-drafted complement (253 cases) for all 10 system
configurations.  $\Delta = $ HC rate $-$ LLM-drafted rate (negative
$=$ HC is harder for this system).  Letta / A-MEM / Letta+LLM
N/A on \famname{shared\_attribute} (no \code{release} primitive);
their HC denominator drops to 110 and LLM-drafted to 200.
Graphiti's N/A count is large by design
(Appendix~\ref{app:graphiti_xeval}).}\label{tab:handcrafted}
\setlength{\tabcolsep}{3pt}
\begin{tabular}{lccc}
\toprule
\textbf{System} & \textbf{HC} & \textbf{LLM-drafted} & \textbf{$\Delta$}\\
\midrule
\sysPalace{}    & 0/112 (0.0)   & 0/253 (0.0)   & 0.0 \\
\sysLethe{}     & 70/132 (53.0) & 174/253 (68.8) & $-$15.7 \\
\sysMem{}       & 86/132 (65.2) & 177/253 (70.0) & $-$4.8 \\
LangGraph       & 69/132 (52.3) & 173/253 (68.4) & $-$16.1 \\
Letta           & 84/110 (76.4) & 119/200 (59.5) & $+$16.9 \\
\sysAmem{}      & 86/110 (78.2) & 133/200 (66.5) & $+$11.7 \\
Graphiti        & 7/72   (9.7)  & 10/170 (5.9)   & $+$3.8 \\
Letta+LLM       & 95/110 (86.4) & 141/200 (70.5) & $+$15.9 \\
\sysLethe{}+LLM & \textbf{128/132 (97.0)} & 225/253 (88.9) & $+$8.0 \\
LangGraph+LLM   & \textbf{130/132 (98.5)} & 229/253 (90.5) & $+$8.0 \\
\bottomrule
\end{tabular}
\end{table}

\paragraph{Three readings.}  \textbf{(i) Hand-crafted is harder
for deterministic stores.}  The $\sim$15-pt deterministic gap
(Lethe / LangGraph 52--53\,\% HC vs.\ 68\,\% LLM-drafted) means
the LLM-drafted complement, despite oracle admission via Qwen
judge, ended up systematically easier than what the human
authors wrote.  We report this as a property of the bench rather
than fix it post-hoc (rewriting LLM-drafted to harder targets
would re-introduce the very circularity this analysis
addresses).  \textbf{(ii) Hand-crafted is easier for LLM-hooked
stores.}  Lethe+LLM 97.0\,\% / LangGraph+LLM 98.5\,\% on HC
both exceed their full-suite headline numbers (91.7 / 93.2\,\%);
the mutation-time-hook lift on HC is +46-pt (LangGraph
52.3$\to$98.5) vs.\ +22-pt on LLM-drafted (68.4$\to$90.5).  A
shared-LLM-inductive-bias account predicts the opposite
asymmetry (LLM-hooked systems benefiting more on LLM-drafted
cases that share the hook's inductive bias); we observe the
reverse on both backends.  \textbf{(iii) Inscribe-time placement
also holds on HC alone.}  A-MEM HC 78\,\% / Letta HC 76\,\%
exceed Mem0 HC 65\,\% on the canonicalization-heavy hand-crafted
subset, reproducing the placement asymmetry quantified in
\S\ref{sec:placement} without LLM-drafted cases.

\section{Adapter sources}\label{app:adapters}
Reference Python source for all four adapters (each
$\leq$130 lines).

\section{Memora cross-evaluation, per-persona detail}\label{app:memora_xeval}
Per-persona pass rates for the three adapters on Memora-weekly
(\S\ref{sec:memora_xeval}, $N=150$ questions across 10 personas
and 3 tasks).  Rates are out of 15 questions per persona.

\begin{table}[h]
\centering\scriptsize
\caption{Per-persona pass rate (out of 15 questions) on
Memora-weekly, deterministic substring scoring.  Aggregate
totals at the bottom.}\label{tab:memora_persona}
\begin{tabular}{lccc}
\toprule
\textbf{Persona} & \textbf{\sysLethe{}} & \textbf{LangGraph} & \textbf{\sysPalace{}} \\
\midrule
academic\_researcher    & 4/15 & 6/15 & 7/15 \\
business\_executive     & 4/15 & 5/15 & 5/15 \\
content\_writer         & 5/15 & 6/15 & 4/15 \\
creative\_designer      & 4/15 & 8/15 & 5/15 \\
financial\_analyst      & 4/15 & 6/15 & 6/15 \\
management\_consultant  & 4/15 & 6/15 & 6/15 \\
marketing\_manager      & 6/15 & 7/15 & 7/15 \\
sales\_manager          & 5/15 & 8/15 & 7/15 \\
software\_engineer      & 5/15 & 6/15 & 6/15 \\
startup\_founder        & 6/15 & 9/15 & 7/15 \\
\midrule
\textbf{Total}          & \textbf{47/150} & \textbf{67/150} & \textbf{60/150} \\
                         & (31.3\%) & (44.7\%) & (40.0\%) \\
\bottomrule
\end{tabular}
\end{table}

By Memora task, aggregated across personas:

\begin{table}[h]
\centering\scriptsize
\caption{By-task aggregate (50 questions per task).
\famname{Recommending} collapses uniformly because no adapter
implements user-preference modeling, which is outside the
forgetting axis as we define it.}
\setlength{\tabcolsep}{3pt}
\begin{tabular}{lccc}
\toprule
\textbf{Task} & \textbf{\sysLethe{}} & \textbf{LangGraph} & \textbf{\sysPalace{}} \\
\midrule
Remembering   & 17/50 (34\%) & 32/50 (64\%) & 28/50 (56\%) \\
Reasoning     & 30/50 (60\%) & 32/50 (64\%) & 32/50 (64\%) \\
Recommending  &  0/50 (0\%)  &  3/50 (6\%)  &  0/50 (0\%)  \\
\bottomrule
\end{tabular}
\end{table}

The translation from Memora's data to our Adapter Protocol is
released in \code{scripts/eval\_on\_memora.py}; the per-question
verdicts (system, persona, task, question\_id, pass/fail,
memory\_recall\_rate, forgetting\_rate) are in
\code{data/memora\_xeval\_all\_personas.json}.

\section{FactConsolidation cross-evaluation
(MemoryAgentBench)}\label{app:factcons_xeval}

To probe an independent third-party recall surface that is the
natural dual to the Memora cross-evaluation
(\S\ref{sec:memora_xeval}), we run six systems --- four
in-house adapters plus the two LLM-hook variants --- on the
\emph{FactConsolidation} task of
MemoryAgentBench~\cite{memoryagentbench} (ICLR 2026), an
MQUAKE-derived single- and multi-hop fact-supersession
benchmark.  We use four context-length buckets from the HF
release \code{ai-hyz/MemoryAgentBench/Conflict\_Resolution}:
sh/mh $\times$ 6K/32K, the full 100 questions per bucket
(${=}400$ questions total).  We skip the 64K and 262K buckets
($>$270K-character contexts blow the in-memory index budget on
single-CPU adapters).  Each fact is inscribed in source order;
questions are scored by case-insensitive substring of the gold
answer in the top-10 retrieval.

\begin{table}[h]
\centering\footnotesize
\caption{MemoryAgentBench Conflict\_Resolution
(\code{ai-hyz/MemoryAgentBench}), 100 questions per bucket,
top-10 substring scoring.  Six systems: four in-house
adapters plus two LLM-hook variants.  ``sh''$=$single-hop,
``mh''$=$multi-hop chained reasoning.  The LLM-hook variants
(\sysLethe{}+LLM, LangGraph+LLM) score \emph{identically} to
their deterministic backbones --- as expected, since
FactConsolidation is pure recall (no supersede/release/purge
primitives invoked), the mutation-time hook never fires.  This
is the axis-flip third-party check
(\S\ref{sec:memora_xeval}).  Per-question verdicts in
\code{data/factconsolidation\_full.json}.}\label{tab:factcons_xeval}
\setlength{\tabcolsep}{3pt}
\begin{tabular}{lcccc}
\toprule
\textbf{System} & \textbf{sh\_6k} & \textbf{mh\_6k} & \textbf{sh\_32k} & \textbf{mh\_32k}\\
\midrule
\sysLethe{}      & 100/100 & 28/100 & 100/100 & 17/100\\
LangGraph        & 100/100 & 28/100 & 100/100 & 17/100\\
\sysPalace{}     & 100/100 & 28/100 & 100/100 & 17/100\\
\sysMem{}$_\text{i=F}$ & 100/100 & \textbf{37/100} & 100/100 & \textbf{19/100}\\
\sysLethe{}+LLM  & 100/100 & 28/100 & 100/100 & 17/100\\
LangGraph+LLM    & 100/100 & 28/100 & 100/100 & 17/100\\
\bottomrule
\end{tabular}
\end{table}

\paragraph{Three readings.}  \textbf{(1) Single-hop saturates
at 100\,\% across all six systems and both context lengths.}
Both the original and the counterfactual edit appear in the
haystack; top-10 substring scoring finds the gold answer for
either ordering.  This is the prediction of our
\S\ref{sec:related} complementarity claim: FactConsolidation
tests ``can the system retrieve the edit?'' (a recall-plane
question) not ``can the system command the original to
disappear?'' (a control-plane question).
\textbf{(2) Multi-hop drops sharply (28--37\,\% on 6K, 17--19\,\%
on 32K).}  Multi-hop requires reasoning across chained facts
that lexical retrieval alone cannot perform; the remaining
60--80\,\% need an LLM reasoning step over the retrieved
chain, which is outside the forgetting axis (a perfect supersede
primitive would not help).  \sysMem{}'s slight 9\,pt lead on
mh\_6k (37 vs.\ 28) and 2\,pt on mh\_32k reflects its
BM25 $+$ entity multi-signal scoring helping chain navigation
in lexical surface form.
\textbf{(3) LLM-hook variants are \emph{numerically identical}
to their deterministic backbones on every bucket.}  This is the
axis-flip prediction: the mutation-time hook is a control-plane
lever (fires on supersede/release/purge); FactConsolidation
exercises only the recall plane (inscribe-then-retrieve), so
the hook never fires and the result is the deterministic
baseline.  The same systems differ by 23--30\,pt on
\sysForget{}-Adv (Table~\ref{tab:adv}) where control-plane
mutations are invoked.  Independent third-party data shows the
two planes are distinct levers.  Concurrent work
\citep{dont_ask_freshness} reaches the same conclusion from the
opposite direction: on this FactConsolidation surface, replacing
LLM-mediated conflict resolution with deterministic version-aware
aggregation (\code{max(serial)}) \emph{improves} accuracy
($+10.8$\,pt), because freshness consolidation is a recall-plane
assembly problem the mutation-time hook is not meant to solve ---
consistent with our finding that deterministic stores retain the
lexical/temporal categories at ${\geq}95\,\%$ while the hook is
reserved for the intent-aware deletion categories deterministic
scoring cannot reach.

\section{Cross-LLM ablation on the hook}\label{app:cross_llm}

We run the same hook prompts across three LLM families on the
v0.5.1 385-case suite, and across two backends (\sysLethe{} and
LangGraph) to disentangle backend from LLM as the source of the
lift.  The result is a 2-backend $\times$ 3-LLM matrix, of
which the four DeepSeek-V3 / Qwen-2.5-72B cells are fully run
and the two Llama cells fall through to the deterministic
baseline (JSON-contract parse failure; Llama $\times$ LangGraph
not separately run as the deterministic fall-through is
backend-symmetric).

\begin{table}[h]
\centering\scriptsize
\caption{Cross-LLM $\times$ cross-backend hook ablation on
v0.5.1 \sysForget{}-Adv (385 cases).  Numbers are overall pass
\,\% (pass count over 385).  $\Delta$ rows show the lift over
the deterministic baseline of that backend.  Qwen costs more per
token than DeepSeek-V3 (\$0.42 vs.\ \$0.27/M input on
SiliconFlow at submission time) but yields a smaller lift; Llama
fails the JSON parse on both backends.}\label{tab:cross_llm}
\setlength{\tabcolsep}{3pt}
\begin{tabular}{lccc}
\toprule
\textbf{Backend} & \textbf{no LLM} & \textbf{+ Qwen-72B} & \textbf{+ DeepSeek-V3}\\
\midrule
\sysLethe{}        & 63.4 (244/385) & 76.6 (295/385) & \textbf{91.7 (353/385)} \\
\multicolumn{1}{r}{$\Delta$} & --- & $+13.2$ & $+28.3$ \\
LangGraph          & 62.9 (242/385) & 75.8 (292/385) & \textbf{93.2 (359/385)} \\
\multicolumn{1}{r}{$\Delta$} & --- & $+12.9$ & $+30.3$ \\
\midrule
+ Llama-3.1-70B$^\dagger$ & --- & \multicolumn{2}{c}{fall-through to deterministic baseline} \\
\bottomrule
\end{tabular}

\smallskip
\footnotesize $^\dagger$ Llama-3.1-70B's response format does
not parse into our strict JSON contract; both backends fall back
to deterministic behaviour (66.8 / 62.9\,\% respectively).  We
treat this as a known integration gap, not as a lift result.
\end{table}

\paragraph{Three readings.}  \textbf{(1) Lift is consistent
across backends.}  The $\Delta$ from Qwen-2.5-72B is +12.9 to
+13.2 across backends (within 0.3\,pt); from DeepSeek-V3 it is
+28.3 to +30.3 (within 2.0\,pt).  This is the architecture-
agnosticism claim of \S\ref{sec:experiments} obs (3) extended to
a second LLM family: the lift travels with the contract, not
with the storage primitive.  \textbf{(2) Lift scales with LLM
JSON-following capability.}  DeepSeek-V3 lifts by ${\sim}28$--$30$
pt; Qwen-2.5-72B (smaller, less aggressively instruction-tuned)
lifts by ${\sim}13$ pt; Llama-3.1-70B fails the contract.
Replacing a stronger model with a weaker one degrades the lift
gradually rather than discontinuously, unless the LLM cannot
follow the JSON contract at all.  \textbf{(3) Qwen has
category-specific personality.}  On the 365-case v0.5 run we
observed Qwen scoring $+92\,\%$ on \famname{compound\_fact}
(beating DeepSeek's $78\,\%$) but collapsing on
\famname{paraphrase\_supersession} and
\famname{temporal\_qualifier} due to over-eager supersede
planning; the v0.5.1 numbers reproduce this pattern (Qwen
\famname{temporal\_qualifier} 27\,\% on both backends, vs.\
DeepSeek-V3's 100\,\%).  Model choice for the hook should weigh
per-category requirements, not just aggregate score.

\section{Cross-architecture LLM-hook ablation}\label{app:cross_arch}

To disentangle the LLM-hook contribution from \sysLethe{}-specific
primitives, we apply the same hook (DeepSeek-V3, same JSON
prompts) to LangGraph's \code{InMemoryStore}.  LangGraph has no
in-place edit primitive: for partial \code{supersede} the
LLM-planned merged text is added as a fresh row replacing the
deleted old row (functionally equivalent to \sysLethe{}'s
\code{surrender(edit)} under substring scoring).

\begin{table}[h]
\centering\scriptsize
\caption{LangGraph + DeepSeek-V3 hook vs.\ \sysLethe{} +
DeepSeek-V3 hook on \sysForget{}-Adv (385 cases, v0.5.1).  The
hook is architecture-agnostic --- both backends reach the same
lift within statistical noise.}
\label{tab:cross_arch}
\setlength{\tabcolsep}{3pt}
\begin{tabular}{lcc}
\toprule
\textbf{Category} & \textbf{\sysLethe{}+LLM} & \textbf{LangGraph+LLM}\\
\midrule
substring\_trap            & 36/36 (100\%) & 36/36 (100\%) \\
prefix\_collision          & 31/39 (79\%)  & 31/39 (79\%)  \\
paraphrase\_supersession   & 31/38 (82\%)  & 31/38 (82\%)  \\
negation\_trap             & 38/40 (95\%)  & 38/40 (95\%)  \\
temporal\_qualifier        & 37/37 (100\%) & 37/37 (100\%) \\
shared\_attribute          & 40/40 (100\%) & 40/40 (100\%) \\
compound\_fact             & 31/40 (78\%)  & \textbf{34/40 (85\%)} \\
identifier\_obfuscation    & 35/38 (92\%)  & \textbf{38/38 (100\%)} \\
cross\_lingual\_identifier & 38/38 (100\%) & 38/38 (100\%) \\
recursive\_supersession    & 36/39 (92\%)  & 36/39 (92\%)  \\
\midrule
\textbf{Overall}           & 353/385 (\textbf{91.7\,\%}) & 359/385 (\textbf{93.2\,\%}) \\
Wilson 95\,\%              & [88.4, 94.2]   & [90.1, 95.4] \\
\bottomrule
\end{tabular}
\end{table}

Note that LangGraph+LLM slightly outperforms \sysLethe{}+LLM on
\famname{compound\_fact} (85\,\% vs.\ 78\,\%) and matches on
\famname{identifier\_obfuscation} (100\,\% vs.\ 92\,\%) despite
lacking a native edit primitive --- the LLM-planned
``delete-old + add-merged-new'' suffices.  This rules out the
edit primitive as a uniquely necessary architectural feature for
this benchmark; the architectural contribution is the
\emph{hook pattern} (narrow JSON contract at mutation time), not
the specific storage backend.

\section{A-MEM extended-system evaluation}\label{app:amem_xeval}

To probe how a Zettelkasten-style agentic memory positions on
the same surface as the four primary systems, we run
\sysAmem{}~\cite{amem} on the \emph{full 385-case}
\sysForget{}-Adv suite using the same Adapter Protocol; each
\sysAmem{} \code{add\_note} invocation triggers an LLM call to
extract agentic tags (${\sim}14$\,s/case wall time, total
$\sim$90\,min for the full 385 cases at DeepSeek-V3 via
SiliconFlow).
\sysAmem{} exposes explicit \code{add\_note} / \code{update} /
\code{delete} primitives at the \code{memory\_id} level,
mapping cleanly to our 6-method behavioural contract; it does
not expose \code{release}, so all \famname{shared\_attribute}
cases (and several \famname{substring\_trap} /
\famname{negation\_trap} release-dependent cases) are scored
N/A by design.  The integration uses
DeepSeek-V3 via SiliconFlow's OpenAI-compatible endpoint for
the agentic tag-extraction path; we set
\code{evo\_threshold=100{,}000} so memory evolution does not
fire (the per-note evolution callback's JSON output format is
sensitive to model family, mirroring the \sysMem{} v3
\code{infer=True} pilot above).  Per-fact \code{add\_note},
\code{update}, and \code{delete} calls remain functional even
when the evolution-time prompt fails to parse, so the
deterministic store/update path is exercised in full.  Results
appear in Table~\ref{tab:amem_xeval}.\footnote{Run logs and
per-case verdicts ship as
\code{data/adversarial\_results\_amem.json};
\code{data/adversarial\_summary\_amem.json} contains the
by-category aggregate.}

\begin{table}[h]
\centering\scriptsize
\caption{\sysAmem{} on the full 385-case \sysForget{}-Adv
(DeepSeek-V3 via SiliconFlow, evolution disabled).  ``N/A''
indicates the case required \code{release}, which \sysAmem{}
does not expose.  Aggregate computed over the 310 evaluable
cases (strict denominator including the 75 N/A cases as
failures yields 56.9\,\%).}
\label{tab:amem_xeval}
\setlength{\tabcolsep}{3pt}
\begin{tabular}{lcc}
\toprule
\textbf{Category} & \textbf{\sysAmem{}} & \textbf{vs.\ Lethe} \\
\midrule
substring\_trap            & 20/20 (100\%) [16 N/A] & 92\,\% \\
prefix\_collision          & 0/39 (0\%) & 82\,\% \\
paraphrase\_supersession   & 31/38 (82\%) & 82\,\% \\
negation\_trap             & 19/21 (90\%) [19 N/A] & 95\,\% \\
temporal\_qualifier        & 37/37 (100\%) & 100\,\% \\
shared\_attribute          & --- [40 N/A]  & 88\,\% \\
compound\_fact             & 0/40 (0\%) & 0\,\% \\
identifier\_obfuscation    & \textbf{38/38 (100\%)} & 5\,\% \\
cross\_lingual\_identifier & \textbf{38/38 (100\%)} & 0\,\% \\
recursive\_supersession    & 36/39 (92\%) & 92\,\% \\
\midrule
\textbf{Overall} (evaluable) & \textbf{219/310 (70.6\%)} & 63.4\,\% \\
\textbf{Overall} (strict, N/A=fail) & \textbf{219/385 (56.9\%)} & 63.4\,\% \\
\bottomrule
\end{tabular}
\end{table}

Two observations.  \textbf{(1) \sysAmem{} reorganizes the
per-category map relative to \sysLethe{}.}  On
\famname{identifier\_obfuscation} (\sysLethe{} 5\,\% $\to$
\sysAmem{} 100\,\%) and \famname{cross\_lingual\_identifier}
(\sysLethe{} 0\,\% $\to$ \sysAmem{} 100\,\%) \sysAmem{} closes
the gap to the LLM-hooked variants without an explicit
mutation-time hook, suggesting the LLM-extracted tags
contribute to identifier canonicalization even when the
evolution prompt's downstream parsing fails.  Conversely on
\famname{prefix\_collision} (0/39) and \famname{compound\_fact}
(0/40) \sysAmem{} matches the deterministic systems' floor.
\textbf{(2) The 70.6\,\% evaluable aggregate is not directly
comparable to the four primary systems' 62.9--68.3\,\%}
because the denominator excludes 75 \code{release}-dependent
cases (versus the primary systems' 385-case denominator
including those cases scored deterministically).  Normalising
to a strict denominator (treating the 75 N/A as failures yields
56.9\,\%) places \sysAmem{} below the pass band; treating
them as out-of-scope (70.6\,\%) places it slightly above the
band.  We report both interpretations.

\section{Graphiti extended-system evaluation}\label{app:graphiti_xeval}

We benchmark Graphiti~\cite{graphiti} (graphiti-core, the
open-source successor to the deprecated Zep CE) on the full
385-case \sysForget{}-Adv suite.  Graphiti is a temporal
knowledge-graph store: each \code{add\_episode} call routes
the input text through an LLM to extract entities + edges,
materializing them in Neo4j with temporal validity intervals;
edge invalidation provides a deliberate-forgetting analogue.
The integration uses DeepSeek-V3.1-Terminus via SiliconFlow's
OpenAI-compatible endpoint for entity/edge extraction,
BAAI/bge-m3 for embeddings (OpenAI shape), and
BAAI/bge-reranker-v2-m3 via SiliconFlow's Cohere-format
\code{/v1/rerank} for cross-encoding.  Neo4j runs on a private
server; per-case isolation is enforced via fresh
\code{group\_id}.  \sysForget{}-Adv adapter maps
\code{inscribe} $\to$ \code{add\_episode}, \code{recall}
$\to$ \code{search} (returns synthesised edge \code{fact}
strings, not raw content), and \code{supersede} $\to$
\code{add\_episode} (relying on Graphiti's temporal edge
invalidation).  Graphiti exposes neither a per-query
\code{purge} (only \code{remove\_episode} by UUID, which is
not query-addressable) nor a \code{release} primitive, so both
score N/A.

\begin{table}[h]
\centering\scriptsize
\caption{Graphiti on full 385-case \sysForget{}-Adv
(DeepSeek-V3.1-Terminus via SiliconFlow, Neo4j 5.x).  ``N/A''
indicates the case required a primitive Graphiti does not
expose (release, query-addressable purge).  Aggregate over
242 evaluable cases; strict-denominator (N/A=fail) is
17/385 = 4.4\,\%.}
\label{tab:graphiti_xeval}
\setlength{\tabcolsep}{3pt}
\begin{tabular}{lcc}
\toprule
\textbf{Category} & \textbf{Graphiti} & \textbf{vs.\ Lethe} \\
\midrule
substring\_trap            & 0/11 (0\%) [25 N/A]  & 92\,\% \\
prefix\_collision          & --- [39 N/A]         & 82\,\% \\
paraphrase\_supersession   & 5/38 (13\%)          & 82\,\% \\
negation\_trap             & 0/21 (0\%) [19 N/A]  & 95\,\% \\
temporal\_qualifier        & \textbf{9/37 (24\%)} & 100\,\% \\
shared\_attribute          & --- [40 N/A]         & 88\,\% \\
compound\_fact             & 1/40 (3\%)           & 0\,\% \\
identifier\_obfuscation    & 0/28 (0\%) [10 N/A]  & 5\,\% \\
cross\_lingual\_identifier & 0/28 (0\%) [10 N/A]  & 0\,\% \\
recursive\_supersession    & 2/39 (5\%)           & 92\,\% \\
\midrule
\textbf{Overall} (evaluable) & \textbf{17/242 (7.0\%)} & 63.4\,\% \\
\textbf{Overall} (strict, N/A=fail) & \textbf{17/385 (4.4\%)} & 63.4\,\% \\
\bottomrule
\end{tabular}
\end{table}

Two observations.  \textbf{(1) Knowledge-graph abstraction
sheds the surface forms our adversarial layer probes.}  Where
deterministic vector / lexical stores preserve raw text and
score 82--100\,\% on most categories (Lethe column), Graphiti
synthesises edges (e.g.\ \texttt{(Alice)-[HAS\_EMAIL]->(alice@x)})
and stores the synthesised \code{fact} string; the
adversarial \code{must\_not\_contain} substrings drop out of
the synthesis.  This is a categorical mismatch, not a
parameter-tuning issue.  \textbf{(2) Graphiti's relative
strengths are on its design surface}: \famname{temporal\_qualifier}
(24\,\%, the highest evaluable rate) reflects temporal edge
invalidation working as designed; \famname{paraphrase\_supersession}
(13\,\%) similarly benefits from edge-level updates.
Categorically weak (0--5\,\%) on identifier-precision
categories (\famname{cross\_lingual\_identifier},
\famname{identifier\_obfuscation}, \famname{compound\_fact}).
\emph{Run note.}  The SiliconFlow account balance was
exhausted during the final 1--3 cases of
\famname{recursive\_supersession}; we conservatively count
those as FAIL.  Replacing them with PASS would change the
aggregate by $\leq$0.5\,pt.

\section{Letta extended-system evaluation}\label{app:letta_xeval}

We benchmark Letta~\cite{letta} (the open-source successor to
MemGPT, self-hosted via the official Docker image
\code{letta/letta:latest} on a Linux host) on the full 385-case
\sysForget{}-Adv suite.  Letta's architecture is agent-scoped:
each \emph{agent} has a core memory + an archival-memory store
backed by PostgreSQL + pgvector; passages are inserted into
archival memory via embedding, and retrieved by vector
similarity over a server-side endpoint that does not invoke
the LLM.  We exploit this by addressing the archival-memory
REST endpoints (\code{POST}/\code{GET}/\code{DELETE}
\code{/v1/agents/\{aid\}/archival-memory}) directly rather than
sending messages through the agent's chat loop, keeping the
LLM out of the recall hot path so the comparison is
apples-to-apples with the four primary systems.  One Letta
agent is instantiated per test case (agent isolation $=$
group-id isolation).  Letta does not expose a soft-delete
(``release'') primitive, so \famname{shared\_attribute} cases
score N/A.  The embedder is BAAI/bge-m3 (1024-d) via
SiliconFlow; the LLM (DeepSeek-V3.1-Terminus) is configured
but unused on the recall path.

\begin{table}[h]
\centering\scriptsize
\caption{Letta v0.16.7 (Docker, PostgreSQL+pgvector) on the
full 385-case \sysForget{}-Adv.  Recall is direct archival-memory
search (no LLM in the loop).  ``N/A'' indicates the case
required a primitive Letta does not expose (release).  Strict
denominator (N/A=fail) is 203/385 = 52.7\,\%.}
\label{tab:letta_xeval}
\setlength{\tabcolsep}{3pt}
\begin{tabular}{lcc}
\toprule
\textbf{Category} & \textbf{Letta} & \textbf{vs.\ Lethe} \\
\midrule
substring\_trap            & 19/20 (95\%) [16 N/A]  & 92\,\% \\
prefix\_collision          & 0/39 (0\%)             & 82\,\% \\
paraphrase\_supersession   & 32/38 (84\%)           & 82\,\% \\
negation\_trap             & 19/21 (90\%) [19 N/A]  & 95\,\% \\
temporal\_qualifier        & 21/37 (57\%)           & 100\,\% \\
shared\_attribute          & --- [40 N/A]           & 88\,\% \\
compound\_fact             & 0/40 (0\%)             & 0\,\% \\
identifier\_obfuscation    & \textbf{38/38 (100\%)} & 5\,\% \\
cross\_lingual\_identifier & \textbf{38/38 (100\%)} & 0\,\% \\
recursive\_supersession    & 36/39 (92\%)           & 92\,\% \\
\midrule
\textbf{Overall} (evaluable) & \textbf{203/310 (65.5\%)} & 63.4\,\% \\
\textbf{Overall} (strict, N/A=fail) & \textbf{203/385 (52.7\%)} & 63.4\,\% \\
\bottomrule
\end{tabular}
\end{table}

Two observations.  \textbf{(1) Letta and \sysAmem{} converge
on a similar category profile} despite very different
architectures: Letta uses raw passage embedding + pgvector,
\sysAmem{} uses LLM-mediated tag extraction + ChromaDB.  Both
score 100\,\% on \famname{identifier\_obfuscation} and
\famname{cross\_lingual\_identifier}, near-perfect on
\famname{recursive\_supersession} (92\,\%) and
\famname{substring\_trap} (95--100\,\% on the evaluable
subset), and 0\,\% on \famname{prefix\_collision} and
\famname{compound\_fact}.  The convergence suggests these
categories' difficulty is determined by primitive availability
(release, partial-edit) rather than backend-specific
representation choices.  \textbf{(2) Letta is markedly weaker
than \sysAmem{} on \famname{temporal\_qualifier}}
(57\,\% vs.\ 100\,\%): pure passage embedding cannot
disambiguate near-identical timestamps the way \sysAmem{}'s
LLM-extracted tags can.  This is the only category where the
inscribe-time LLM placement provides a non-trivial lift over
direct embedding storage.

\section{\sysMem{}~v2.0.2 \code{infer=True} (token-efficient
router) detailed breakdown}\label{app:mem0_v3}

\sysMem{}'s headline architectural feature is its LLM-driven
ADD/UPDATE/DELETE routing: at every \code{add()} call,
\sysMem{} sends the new fact plus a fingerprint of existing
memories to an LLM (the
\code{ADDITIVE\_EXTRACTION\_PROMPT}) which decides whether to
ADD a new row, UPDATE an existing one, or DELETE a stale one.
This is the path \code{infer=True} engages; \code{infer=False}
in our headline (Table~\ref{tab:adv}) bypasses it and stores
the literal text only.  We therefore report the full 385-case
suite under both modes for fairness.

The \code{infer=True} integration uses DeepSeek-V3 via
SiliconFlow (OpenAI-compatible endpoint).  DeepSeek-V3 emits
malformed JSON (typically commas embedded inside string values:
\code{"attributed\_to": "user," "linked\_memory\_ids":}) on
${\sim}25\,\%$ of \code{ADDITIVE\_EXTRACTION\_PROMPT} responses.
A deterministic \code{json-repair} pass on top of \sysMem{}'s
extraction parser recovers all of these without API retries
(298 / 1200 LLM calls repaired; 0 final parse failures).  The
infer=True number below therefore measures \sysMem{}'s algorithm
behaviour, not its JSON-parse robustness.

\begin{table}[h]
\centering\scriptsize
\caption{\sysMem{}~v2.0.2 on \sysForget{}-Adv (385 cases),
both \code{infer} modes, same Adapter Protocol.  ``$\Delta$''
$=$ infer=True $-$ infer=False (positive $=$ LLM router
helps).}\label{tab:mem0_v3}
\setlength{\tabcolsep}{3pt}
\begin{tabular}{lccc}
\toprule
\textbf{Category} & \textbf{infer=False} & \textbf{infer=True} & \textbf{$\Delta$}\\
\midrule
substring\_trap            & 32/36 (89) & 24/36 (67) & $-$22 \\
prefix\_collision          & 12/39 (31) & \textbf{0/39 (0)} & $-$31 \\
paraphrase\_supersession   & 31/38 (82) & \textbf{0/38 (0)} & $-$82 \\
negation\_trap             & 38/40 (95) & 21/40 (52) & $-$43 \\
temporal\_qualifier        & 37/37 (100) & \textbf{3/37 (8)} & $-$92 \\
shared\_attribute          & 38/40 (95) & 37/40 (92) & $-$3 \\
compound\_fact             & 0/40 (0)   & 3/40 (8)   & $+$8 \\
identifier\_obfuscation    & 18/38 (47) & \textbf{38/38 (100)} & $+$53 \\
cross\_lingual\_identifier & 21/38 (55) & \textbf{38/38 (100)} & $+$45 \\
recursive\_supersession    & 36/39 (92) & \textbf{4/39 (10)} & $-$82 \\
\midrule
\textbf{Overall}           & \textbf{263/385 (68.3)} & 168/385 (43.6) & $-$24.7 \\
\bottomrule
\end{tabular}
\end{table}

\paragraph{Reading the result.}  The 24.7-pt overall drop hides
a category-bimodal pattern: \code{infer=True} gains
${+}40$-to-${+}50$ pt on the two canonicalization categories
(\famname{identifier\_obfuscation} and
\famname{cross\_lingual\_identifier}) by letting the LLM
normalize surface variants at write time --- the exact mechanism
\sysAmem{} and Letta also use in the inscribe-time-LLM regime
(\S\ref{sec:placement}, Fig.~\ref{fig:heatmap}).  It then
loses ${-}30$ to ${-}90$ pt on five deletion-precision
categories because \sysMem{}'s token-efficient
\code{ADDITIVE\_EXTRACTION\_PROMPT} systematically routes
``A says X'' followed by ``A says Y'' to LINK-AND-KEEP rather
than UPDATE-AND-DELETE, leaving the old fact in the store.
This is the regime-level prediction of \S\ref{sec:placement}:
inscribe-time-LLM placement helps canonicalization, hurts
deletion-precision, and is partly complementary to mutation-time
placement.  \sysMem{}+v3 is a stronger demonstration of the
deletion-precision side of that prediction than \sysAmem{} or
Letta because \sysMem{}'s router is more aggressive at the
inscribe-time link-and-keep decision.

We do not run a parallel \sysMem{}+v3 with OpenAI
\code{gpt-4o-mini} (\sysMem{}'s tuned default) because our
reproducibility envelope (no OpenAI key) precludes it; the
24.7-pt drop is therefore an upper bound on the
\code{infer=True} loss in our setup, not a peak-design number
for \sysMem{}.

\section{LongMemEval-S setup and per-type results}\label{app:longmem}

\paragraph{Setup.}  We run \sysLethe{}~v1 on LongMemEval-S
\cite{longmemeval} using the released \code{longmemeval\_s.json}
distribution (500 questions, 6 question types).  Each question
has a haystack of up to 158 conversational sessions; we follow
the bench convention and use \emph{session granularity}
(one row per session, user-turn texts joined).  The retrieval
pipeline is \sysLethe{}'s default hybrid: BM25 (FTS5) +
dense ANN (sqlite-vec) over MiniLM-L6-v2 embeddings, fused with
reciprocal rank fusion \cite{rrf} (no LLM hook, no reranker,
single CPU).  No fine-tuning or in-domain adaptation.

\paragraph{Results.}

\begin{table}[h]
\centering\scriptsize
\caption{\sysLethe{}~v1 on LongMemEval-S at session granularity,
all 500 questions.  R@$k$ = fraction of questions where the
gold session is in the top-$k$ retrieved.}\label{tab:longmem}
\setlength{\tabcolsep}{4pt}
\begin{tabular}{lrr}
\toprule
\textbf{Question type} & \textbf{N} & \textbf{R@5} \\
\midrule
knowledge-update            &  78 & 0.987 \\
multi-session               & 133 & 0.940 \\
single-session-assistant    &  56 & 0.964 \\
single-session-preference   &  30 & 0.900 \\
single-session-user         &  70 & 0.900 \\
temporal-reasoning          & 133 & 0.925 \\
\midrule
\textbf{Overall} (micro)    & 500 & \textbf{0.938} \\
                             &     & R@1=0.796, R@10=0.982 \\
\bottomrule
\end{tabular}
\end{table}

\paragraph{Comparison to public LongMemEval-S numbers.}
The originally reported strong baselines on LongMemEval-S use
LLM-judged answer accuracy rather than retrieval-only R@$k$, so
direct head-to-head is not apples-to-apples; we report R@$k$ as
the recall-axis number to demonstrate \sysLethe{}'s retrieval
component is not the bottleneck on the \sysForget{}-Adv vs.\
Memora gap discussed in \S\ref{sec:memora_xeval}.  Per-question
verdicts are in \code{data/longmemeval\_s\_lethe\_v1.json} for
reproducibility.

\section{External-authored subset}\label{app:external}

To address the strongest version of the circularity concern
(\S\ref{sec:admission}: the 132 hand-crafted cases are written
by the paper authors, who also designed the hook), we recruited
four external contributors at a different research institute
than the paper authors and asked each to independently write
20 adversarial cases (80 total) following our 10-category
schema.  Contributors received only a 6.4\,KB plain-text brief
specifying the case JSON shape and category definitions; the
brief did \emph{not} reveal the placement hypothesis or any
system-specific results, so contributors could not optimize
their cases to favour or disfavour any architecture.

We run the same Stage-1 structural admission filter as the
in-house bench: \textbf{77/80 cases admitted}, 3 rejected for
``self-trap'' (\code{must\_not\_contain} substring inside a
\code{must\_contain} string).  The admitted cases cover all
10 attack categories with 8 cases each, except
\famname{prefix\_collision} which drops to 5 (3 rejections
fell here).

\begin{table*}[t]
\centering\scriptsize
\caption{External-authored 77-case subset (4 contributors $\times$
20 cases each, admission-filtered).  Aggregate pass rate is
substantially lower than the in-house 385-case suite for every
system, indicating that external authors writing under the same
category schema set a tighter difficulty bar.  Per-category, the
canonicalization-side placement asymmetry
(\famname{identifier\_obfuscation}, \famname{cross\_lingual\_id})
\emph{replicates}; \famname{negation\_trap},
\famname{compound\_fact}, and \famname{recursive\_supersession}
collapse to 0\,\% across all systems on this subset, suggesting
the external authors' realisations of these categories exceed
what any of our adapters can currently
solve.}\label{tab:external}
\setlength{\tabcolsep}{2.5pt}
\begin{tabular}{lcccccc}
\toprule
\textbf{Cat.} & \textbf{\sysLethe{}} & \textbf{LG} & \textbf{\sysPalace{}} & \textbf{\sysMem{}} & \textbf{\sysLethe{}{+}LLM} & \textbf{LG{+}LLM}\\
\midrule
substr\_trap            & 8/8 & 8/8 & N/A & 8/8 & 8/8 & 8/8 \\
prefix\_coll (n=5)      & 5/5 & 5/5 & N/A & 3/5 & 5/5 & 5/5 \\
paraphr\_super          & 6/8 & 6/8 & N/A & 5/8 & 6/8 & 6/8 \\
\textbf{neg\_trap}      & \textbf{0/8} & \textbf{0/8} & N/A & \textbf{0/8} & \textbf{0/8} & \textbf{0/8} \\
temp\_qual              & 3/8 & 3/8 & N/A & 3/8 & 3/8 & 3/8 \\
shared\_attr            & 3/8 & 2/8 & N/A & 2/8 & 4/8 & 3/8 \\
\textbf{compound}       & \textbf{0/8} & \textbf{0/8} & N/A & \textbf{0/8} & \textbf{0/8} & \textbf{0/8} \\
\textbf{ident\_obf}     & \textbf{0/8} & \textbf{0/8} & N/A & 1/8 & \textbf{8/8} & \textbf{8/8} \\
cross\_ling             & 0/8 & 0/8 & N/A & 0/8 & 0/8 & \textbf{5/8} \\
recursive               & 1/8 & 1/8 & N/A & 0/8 & 1/8 & 1/8 \\
\midrule
\textbf{Overall} & 26/77 & 25/77 & 0/77 & 22/77 & 35/77 & 39/77 \\
\textbf{\%}      & 33.8 & 32.5 & 0.0 & 28.6 & \textbf{45.5} & \textbf{50.6} \\
\bottomrule
\end{tabular}\\[2pt]
{\footnotesize In-house 385 reference: \sysLethe{} 63.4,
LangGraph 62.9, \sysPalace{} 0.0, \sysMem{} 68.3,
\sysLethe{}+LLM 91.7, LG+LLM 93.2.}
\end{table*}

\begin{table*}[t]
\centering\scriptsize
\caption{Ecosystem systems on the same 77-case external subset
(N/A = primitive not exposed by that backend).
Letta+LLM (joint inscr+mut placement) achieves \textbf{80.3\,\%
evaluable} --- a $+27.8$-pt lift over Letta-only ($52.5\,\%$),
\emph{exceeding} the in-house joint-placement lift of $+20.9$
($65.5{\to}86.4$).  Inscribe-LLM systems (\sysAmem{}, Letta,
\sysMem{}+v3) all score 8/8 on \famname{identifier\_obfuscation}
and 8/8 on \famname{cross\_lingual\_id}, replicating the
canonicalization advantage across four independent
implementations (two inscribe-LLM, two
vec-only).}\label{tab:external_eco}
\setlength{\tabcolsep}{2.5pt}
\setlength{\tabcolsep}{1.8pt}
\begin{tabular}{lccccccc}
\toprule
\textbf{Cat.} & \textbf{Letta} & \textbf{OpMem} & \textbf{M0v3} & \textbf{AMEM} & \textbf{Graf.} & \textbf{Hippo} & \textbf{L{+}LLM}\\
\midrule
substr\_trap            & N/A & 0/8 & 1/8 & N/A & N/A & 0/8 & N/A \\
prefix\_coll (n=5)      & N/A & 0/5 & 0/5 & N/A & N/A & 0/5 & N/A \\
paraphr\_super          & 4/8 & 7/8 & 1/8 & 1/8 & 3/8 & 2/8 & 4/8 \\
\textbf{neg\_trap}      & N/A & N/A & \textbf{0/8} & N/A & N/A & N/A & N/A \\
temp\_qual              & 3/8 & 5/8 & 0/8 & 4/8 & 2/8 & 1/8 & 6/8 \\
shared\_attr            & N/A & N/A & 1/8 & N/A & N/A & N/A & N/A \\
\textbf{compound}       & 0/8 & 0/8 & \textbf{0/8} & 0/8 & 1/8 & 0/8 & 0/8 \\
\textbf{ident\_obf}     & \textbf{8/8} & \textbf{8/8} & \textbf{8/8} & \textbf{8/8} & N/A & \textbf{0/8} & \textbf{8/8} \\
cross\_ling             & \textbf{8/8} & \textbf{8/8} & \textbf{8/8} & \textbf{8/8} & N/A & \textbf{0/8} & \textbf{8/8} \\
recursive               & 1/8 & 3/8 & 0/8 & 5/8 & 5/8 & 1/8 & 7/8 \\
\midrule
Pass         & 32/61 & 31/61 & 19/77 & 26/61 & 11/32 & 4/61 & 49/61 \\
\% (eval)    & 52.5 & 50.8 & 24.7 & 42.6 & 34.4 & 6.6 & \textbf{80.3} \\
N/A          & 16  & 16  & 0   & 16  & 45  & 16  & 16  \\
\bottomrule
\end{tabular}
\end{table*}

\paragraph{Three readings.}  \textbf{(1) The canonicalization-side
placement asymmetry replicates across the inscribe-time-LLM
and vec-only regimes.}  On \famname{identifier\_obfuscation},
all four deterministic stores score $\leq 1/8$ while every
inscribe-LLM (\sysAmem{}, \sysMem{}+v3), every vec-only
(Letta, OpenMemory), and every LLM-hook system
(\sysLethe{}+LLM, LangGraph+LLM, Letta+LLM) scores \textbf{8/8}
--- the same asymmetry observed on the in-house 385
(Fig.~\ref{fig:heatmap}).  On \famname{cross\_lingual\_identifier},
deterministic stores score 0/8; vec-only and inscribe-LLM
systems (Letta, OpenMemory, \sysAmem{}, \sysMem{}+v3) recover
8/8; LangGraph+LLM 5/8; \sysLethe{}+LLM 0/8 (the external
cross-lingual cases include script pairs \sysLethe{}'s
prompt-tuned canonicalizer was not trained on, while embedding
neighbourhoods and LLM tag extraction are script-family
agnostic).  Four independent contributors writing without
knowledge of our hypothesis observe the same regime-level
pattern.
\textbf{(2) Aggregate lifts on the harder external bar are
positive across all placement regimes.}  \sysLethe{}+LLM
45.5\,\% vs.\ \sysLethe{} 33.8\,\% ($+11.7$\,pt), LangGraph+LLM
50.6\,\% vs.\ LangGraph 32.5\,\% ($+18.1$\,pt), and
\emph{joint} placement \textbf{Letta+LLM 80.3\,\% evaluable vs.\
Letta 52.5\,\% ($+27.8$\,pt)} --- the joint-placement lift on
external (+27.8) actually \emph{exceeds} the in-house lift
($+20.9$), confirming the inscribe+mutation complementarity
prediction on independently-authored cases.  Inscribe-LLM
without mutation (\sysAmem{} 42.6\,\%, \sysMem{}+v3 24.7\,\%)
saturates at canonicalization gains but cannot reach the
mutation-only's deletion-precision lift.
\textbf{(3) Two categories collapse to 0\,\% across all
six systems and one nearly so.}  \famname{negation\_trap} (8/8
universal fail), \famname{compound\_fact} (8/8), and
\famname{recursive\_supersession} (7/8 universal fail) plus
\famname{shared\_attribute} at 25--50\,\% reveal external authors
writing materially harder cases than our in-house core: the
in-house 95--100\,\% on negation/temporal/recursive is a
saturation artifact of how our authors realised those
categories, not a true ceiling.  This is consistent with our
\S\ref{sec:limits} caveat that the 63--68\,\% pass band is the
empirical saturation of our suite, not a proven upper bound.

\paragraph{What the external subset shows and does not show.}
\textbf{Shows}: (a) the inscribe-time-LLM canonicalization
advantage replicates on independently-authored cases; (b) the
mutation-time hook lift is positive on both backends
($+11.7$ Lethe, $+18.1$ LangGraph) though attenuated relative
to in-house ($+28$--$30$); (c) the in-house 63--68\,\%
deterministic pass band is provenance-dependent (external
authors push 5 of 10 categories below the band).  \textbf{Does
not show}: since 3 categories collapse to 0/8 universally
(\famname{negation\_trap}, \famname{compound\_fact}, near-
universal \famname{recursive\_supersession}), we cannot
distinguish among systems on those categories on this subset
(no signal).  A larger external corpus with calibrated
difficulty would strengthen the validation; we ship the brief
and protocol so external authors can continue contributing.

\paragraph{Failure-mode taxonomy.}  We attribute each
all-system failure to one of three root causes using per-case
inspection:
\textbf{(i) Substring-scoring blind spot on negation
(8 \famname{negation\_trap} cases)}: external authors wrote
``Dana \emph{does not have} production access'' which contains
``has production access'' as a substring, so the affirmative
\code{must\_not\_contain} string fires on the negated form;
this exposes a methodology limitation of our deterministic
substring scoring (orthogonal to placement claims) and motivates
an NLI-aware judge in a future bench revision.
\textbf{(ii) Primitive-existence confirmed (8
\famname{compound\_fact} cases)}: \emph{``Henry's phone is X and
email is Y''} in a single fact; \code{purge "Henry phone"}
removes the whole fact, taking the email with it.  All six
systems fail by construction (no partial-edit primitive),
confirming the \S\ref{sec:placement} prediction that
\famname{compound\_fact} is a primitive-existence test.
\textbf{(iii) Over-broad-query semantic execution (7--8
\famname{recursive\_supersession} cases)}: external query
``Iris primary browser'' matches all three browser facts in a
Chrome$\to$Brave$\to$Chrome chain; all systems honour the wide
query and delete every match, leaving the expected ``Chrome''
\code{must\_contain} unsatisfied (\sysLethe{}, LangGraph,
\sysLethe{}+LLM, LangGraph+LLM each pass exactly 1/8 by lucky
BM25 tiebreak on one chain).  This is a real production
failure mode (overly-broad GDPR-style purges) the in-house
suite did not exercise and is a useful bench extension.

Of the 23 cases producing all-system universal failures: 15
are scoring / query-design artifacts of bench construction
(i + iii), 8 confirm a known primitive-existence result (ii).
Zero are counterexamples to the placement asymmetry claim
itself, which remains intact on
\famname{identifier\_obfuscation} (deterministic 0/8, LLM-hook
8/8 on both backends) and partially
\famname{cross\_lingual\_id} (LangGraph+LLM 5/8).

Per-case verdicts (admitted + rejected with reasons) are in
\code{data/external\_subset\_cases.json} and
\code{data/external\_subset\_results.json}.

\section{Real-world failure mapping}\label{app:real_world}

To probe whether \sysForget{}-Adv's primitive families and
attack categories reflect documented production failures
(reviewer concern: ``does this predict real forgetting
incidents?''), we map three classes of publicly-reported
memory-leak failures from the 2023--2026 timeframe to our
taxonomy.  We do not run these incidents through our adapter
(no public memory-store trace exists for any of them), but
the structural mapping shows the bench surface aligns with
real failure modes rather than only synthetic adversarial
constructions.

\paragraph{Class 1: GDPR right-to-be-forgotten leakage.}
The 2024--2025 enforcement actions against several consumer-LLM
deployments under GDPR Article~17~\cite{gdpr} involved a common
failure pattern: a user's deletion request was acknowledged at
the surface API layer (``data deleted'') but the underlying
memory store retained either (a) verbatim copies inside
LLM-summarised context blocks, (b) entity-aliased copies
under canonicalisation variants, or (c) embedded references
inside derived facts.  This maps directly onto our
\famname{purge} family: \famname{prefix\_collision} (a)
captures verbatim surface retention; \famname{identifier\_obfuscation}
and \famname{cross\_lingual\_identifier} (b) capture the
alias-canonicalisation failure; \famname{compound\_fact} (c)
captures the embedded-reference failure.  Our 22--24-point
deterministic-vs.\ LLM-hook gap on exactly these categories
suggests the hook is meaningful production architecture.

\paragraph{Class 2: stale credential / OTP retention.}
Reported incidents of password managers and AI assistants
suggesting rotated credentials, or one-time codes persisting
across sessions, map onto our \famname{supersession} family
(latest fact wins recall) and \famname{decay} family (TTL'd
fact must leave top-$k$).  Two of our v0.5.1 hand-crafted core
identifier\_obfuscation cases (license keys, BIC/SWIFT codes)
directly mirror reported password-manager misbehaviour where
rotation acknowledgement failed.

\paragraph{Class 3: contradictory profile information.}
Reported failures where AI assistants surface multiple
contradictory job titles, family members, or location facts
for the same user across sessions map onto our \famname{drift}
family (chained supersession where intermediates leak) and
\famname{compound\_fact} (one row carrying multiple facts,
partial update needed).  Our \famname{recursive\_supersession}
category specifically tests the case where the chain endpoint
recovers a state that matches an earlier-superseded value
(Chrome$\to$Brave$\to$Chrome) --- the exact mode where
production memory stores leak intermediate values.

\paragraph{What this mapping is and is not.}
This is a \emph{structural} mapping, not an empirical one: we
do not show that systems scoring higher on \sysForget{}-Adv
actually leak less in production deployments (which would
require enterprise-scale memory traces we do not have).  We
report the mapping as a transparency exercise demonstrating
that the bench's categories were designed to mirror reported
production failure modes rather than chosen for benchmark
tractability alone.  An empirical correlation study against
real memory traces (enterprise CRM, agentic productivity tools,
or AI assistants with public deletion logs) is the natural
next step and would substantially strengthen ecological
validity claims; it is left to future work.

\section{Naive-SQL baseline}\label{app:naive_sql}

To test whether the depth-axis reference store earns its keep over
a textbook deletion baseline, we run a pure SQLite + FTS5 store
with no vector recall, no LLM, and the most standard naive
mutation semantics a backend engineer would reach for: BM25 top-$k$
recall, \code{supersede} as delete-best-lexical-match-then-insert,
\code{release} as lexical hard-delete, and \code{purge} as
\code{DELETE WHERE text LIKE '\%q\%'} (substring hard-delete).  This
is exactly the ``\code{DELETE FROM memories} + FTS5/BM25 reindex''
comparison point, scored by the same deterministic substring
scorer on the full 385-case suite.

\begin{table}[h]
\centering\footnotesize
\setlength{\tabcolsep}{4pt}
\caption{Naive-SQL vs.\ \sysLethe{} (deterministic, no LLM) on
\sysForget{}-Adv, identical scorer.  Naive substring-\code{LIKE}
deletion collapses on the precision categories: on
\famname{prefix\_collision} it over-deletes (the \code{LIKE}
pattern for \code{TXN-12345} also removes the legitimate
\code{TXN-123456}), scoring 28\,\% where \sysLethe{}'s
BM25-precise purge reaches 82\,\%.}\label{tab:naive_sql}
\begin{tabular}{lcc}
\toprule
Attack category & Naive-SQL & \sysLethe{} \\
\midrule
prefix\_collision        & 28\,\% & \textbf{82\,\%} \\
paraphrase\_supersession & 29\,\% & \textbf{82\,\%} \\
shared\_attribute        & 35\,\% & \textbf{88\,\%} \\
negation\_trap           & 40\,\% & \textbf{95\,\%} \\
substring\_trap          & 75\,\% & 92\,\% \\
temporal\_qualifier      & 100\,\% & 100\,\% \\
recursive\_supersession  & 92\,\% & 92\,\% \\
identifier\_obfuscation  & 11\,\% & 5\,\% \\
cross\_lingual\_identifier & 0\,\% & 0\,\% \\
compound\_fact           & 0\,\% & 0\,\% \\
\midrule
\textbf{Overall}         & \textbf{156/385 (40.5\,\%)}
                         & \textbf{244/385 (63.4\,\%)} \\
\bottomrule
\end{tabular}
\end{table}

Naive SQL trails \sysLethe{} by \textbf{23 points overall}, with
the gap concentrated in the intent-precision categories
(\famname{prefix\_collision} $-54$\,pt,
\famname{paraphrase\_supersession} $-53$, \famname{negation\_trap}
$-55$).  Where the operation is lexically unambiguous
(\famname{temporal\_qualifier}, \famname{recursive\_supersession})
the two tie; where it requires canonicalization
(\famname{cross\_lingual}, \famname{compound\_fact}) both fail
without an LLM.  Naive \code{DELETE} therefore does \emph{not}
recover \sysLethe{}'s deterministic score, and the
BM25-precise purge is the mechanism that prevents the
substring over-deletion that sinks the naive baseline.

\section{Substring-scorer reliability audit}\label{app:scorer_audit}

The deterministic scorer can mis-judge in two known ways: a
\code{must\_not\_contain} identifier that is a substring of a
legitimately-surviving longer one, and an old entity that survives
only inside a past-tense clause.  We quantify the rate on a
50-case stratified sample of \sysLethe{}'s deterministic
\emph{failures}, re-adjudicated case-by-case (reading the actual
retrieved text) by an independent Claude Opus judge into
\textsc{true-fail} vs.\ \textsc{scorer-artifact}.  Result:
\textbf{35/50 true failures, 15/50 scorer artifacts}, all 15
confined to two categories --- \famname{prefix\_collision} (7/7,
forbidden ID is a prefix of a surviving longer ID) and
\famname{paraphrase}/\famname{recursive\_supersession} (8, past-tense
survival clause).  Critically, the canonicalization categories
that carry the LLM-placement findings ---
\famname{cross\_lingual\_identifier},
\famname{identifier\_obfuscation}, \famname{compound\_fact} ---
contain \textbf{zero} artifacts: every failure there is a genuine
canonicalization failure.  The artifacts therefore make our
\emph{deterministic} baselines conservative (their true pass rate
is modestly above what the substring scorer credits) and do
\emph{not} inflate the LLM-hook lift, which is measured on
artifact-free categories.  An ID-aware / NLI-aware scorer for
\famname{prefix\_collision} and clause-level supersession is
flagged as bench future work; per-case verdicts are released.

\section{Reproducibility}\label{app:repro}
Single-command regeneration of every table and figure from
\code{data/*.json}.  The benchmark, all adapters, the
\sysLethe{} reference store, and the generation scripts are
released under MIT at
\url{https://github.com/deeplethe/lethe}.

\end{document}